\RequirePackage{fix-cm}
\documentclass[journal]{IEEEtran}
\usepackage{tikz}
\usetikzlibrary{positioning, calc, shadows, shapes, backgrounds, fit}
\definecolor{optblue}{RGB}{0, 114, 189}    
\definecolor{sarorange}{RGB}{217, 83, 25}  
\definecolor{bglight}{RGB}{245, 245, 245}  
\usepackage[export]{adjustbox}
\usepackage{microtype}
\usepackage{amsmath,amsfonts}
\usepackage{algorithmic}
\usepackage{algorithm}
\usepackage{array}
\usepackage[caption=false,font=normalsize,labelfont=sf,textfont=sf]{subfig}
\usepackage{newunicodechar} 
\usepackage{CJKutf8}
\usepackage{textcomp}
\usepackage{stfloats}
\usepackage{url}
\usepackage{verbatim}
\usepackage{graphicx}
\usepackage{booktabs}
\usepackage{rotating}
\usepackage{forest}
\usepackage{makecell}
\usepackage{tabularx}
\usepackage{multirow}
\usepackage{xcolor}
\usepackage{cite} 
\usepackage{amssymb}
\usepackage{pifont}
\usepackage[table]{xcolor}
\newcommand{\cmark}{\ding{51}}%
\newcommand{\xmark}{\ding{55}}%

\usepackage[pagebackref=false, 
            breaklinks=true, 
            colorlinks, 
            citecolor=blue, 
            linkcolor=blue, 
            urlcolor=blue]{hyperref}
\usepackage{tabularray}
\usetikzlibrary{shadows}
\begin{document}


\title{GeoMamba: A Geometry-driven MambaVision Framework and Dataset for Fine-grained Optical-SAR Object Retrieval}

\author{Tiantong Fang, Xiuwei Wang, Jing Xiao, \IEEEmembership{\textit{Senior Member, IEEE}}, Wujie Zhou, \IEEEmembership{\textit{Senior Member, IEEE}}, \\ Liang Liao, \IEEEmembership{\textit{Senior Member, IEEE}}, Mi Wang, \IEEEmembership{\textit{Member, IEEE}}
\thanks{T. Fang, X. Wang and J. Xiao are with the School of Artificial Intelligence, Wuhan University, e-mail: jing@whu.edu.cn.}
\thanks{W. Zhou is with the School of Artificial Intelligence and Information Engineering, Zhejiang University of Science \& Technology (e-mail: wujiezhou@163.com).}
\thanks{L. Liao is with the Hangzhou Institute of Technology, Xidian University, e-mail: liaoliang01@xidian.edu.cn.}
\thanks{M. Wang is with the State Key Laboratory of Information Engineering in Surveying, Mapping and Remote Sensing, Wuhan University, e-mail: wangmi@whu.edu.cn.}}



\maketitle

\begin{abstract}
Multi-source remote sensing enables complementary observation of ground objects, while cross-modal fine-grained object retrieval remains challenging, especially under unaligned optical and SAR conditions. Unlike conventional retrieval settings that rely on paired or spatially aligned samples, practical optical-SAR retrieval is affected by substantial modality discrepancy, speckle noise, and structural inconsistency, which limit robust cross-modal representation learning. To address this problem, we propose GeoMamba, a geometry-driven framework tailored for optical-SAR fine-grained retrieval. Specifically, GeoMamba introduces a Geometric Feature Injection (GFI) module that enhances cross-modal feature interaction and incorporates structural priors, thereby improving the robustness of SAR representations and promoting geometry-consistent feature learning. In addition, a Geometric Consistency Constraint (GCC) module, together with a Deep Supervision (DS) strategy, imposes hierarchical geometric constraints using classical operators, which helps preserve informative object structures during representation learning. We further construct a new dataset, FGOS-as, containing 11 aerospace and maritime categories for evaluating unaligned cross-modal fine-grained object retrieval in realistic remote sensing scenarios. Extensive experiments on FGOS-as demonstrate that GeoMamba outperforms existing methods, achieving 63.3\% mAP and 77.0\% Rank-1 accuracy in all-to-all retrieval setting.
\end{abstract}


\section{Introduction}

\IEEEPARstart{W}{ith} the rapid advancement of Earth observation technologies, multi-source remote sensing imagery has become increasingly accessible, enabling complementary observations of the Earth's surface across heterogeneous modalities such as optical and SAR~\cite{tupin2003detection,wang2025m4,zhao2025towards}. This capability is crucial for a wide range of applications including maritime surveillance~\cite{zhang2020intelligent}, disaster response~\cite{kwak2016disaster}, and environmental monitoring~\cite{yu2022coastline}, where the same object often needs to be identified across different sensing sources. Beyond coarse category-level matching, many real-world scenarios require fine-grained instance discrimination, such as retrieving the same fishing vessel or a specific Boeing 737 aircraft from large-scale cross-modal archives~\cite{zhang2024optical,ahmed2025dual,rane2025machine}, which is essential for reliable object association, persistent monitoring, and cross-source verification. To address this demand, cross-modal fine-grained object retrieval (CM-FGOR) aims to establish robust instance-level correspondence across different remote sensing modalities, thereby supporting precise cross-modal object association and downstream geospatial analysis.

\begin{figure*}[htbp]
    \centering
    \includegraphics[width=0.95\textwidth]{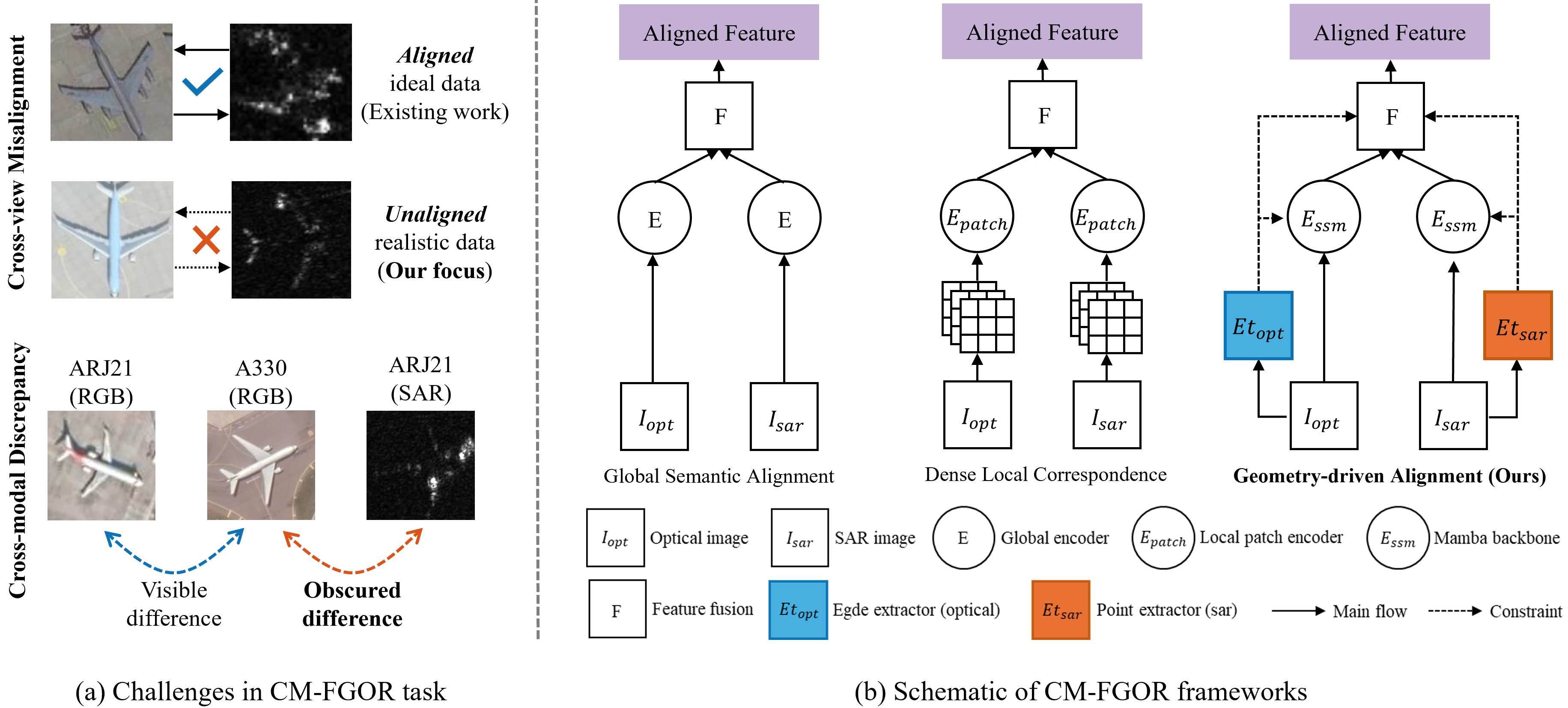} 
    
    \caption{Overview of cross-modal retrieval challenges and our proposed paradigm. \textbf{(a) Challenges in CM-FGOR task:} \textbf{Cross-view Misalignment} highlights our realistic unaligned setting, contrasting with previous works needing strict pixel-level pairs. \textbf{Cross-modal Discrepancy} shows the vast modality gap, where fine-grained optical differences (\emph{e.g.}, ARJ21 vs. A330) are obscured in SAR. \textbf{(b) Schematic of CM-FGOR framework:} \emph{Global Semantic Alignment} maps inputs to a shared space but loses spatial details. \emph{Dense Local Correspondence} uses patch matching but fails under severe misalignment. \emph{Geometry-driven Alignment (Ours)} bridges the modality gap using a dual Mamba backbone ($E_{ssm}$) and structural extractors ($Et_{opt}$, $Et_{sar}$). This allows direct geometric feature injection into the SAR flow and applies consistency constraints on the feature fusion (F) module for final alignment.}
    
    \label{fig:intro_comparison}
\end{figure*}

Existing cross-modal retrieval approaches relevant to CM-FGOR can be broadly categorized into two paradigms: global semantic alignment and dense local correspondence. Global alignment methods project heterogeneous inputs into a shared latent space by optimizing holistic image-level embeddings~\cite{xiong2020deep,sun2021multisensor,huang2024deep}. While effective for coarse category retrieval, these methods tend to compress spatial details into compact global representations, thereby weakening subtle discriminative cues and increasing sensitivity to background clutter. In contrast, dense local correspondence methods attempt to preserve fine-grained characteristics through patch-level or part-based matching across modalities~\cite{yang2025cross,luo2026dynamic}. However, as illustrated in Fig. \ref{fig:intro_comparison}(a), directly applying both paradigms to practical CM-FGOR remains challenging due to two fundamental issues:
\begin{itemize}
    \item \textbf{Cross-view Misalignment:} Most existing methods are developed under the assumption of spatially aligned cross-modal pairs. In practical optical-SAR scenarios, especially in modern multi-constellation sensing settings, images are acquired at different times and viewing geometries, leading to severe spatial and temporal misalignment. 
    \item \textbf{Cross-modal Discrepancy:} Optical-SAR retrieval is further hindered by a huge modality gap. Optical imagery captures continuous surface appearances and texture, whereas SAR imagery is characterized by discrete electromagnetic scattering centers and speckle noise, which severely obscures fine-grained identity cues. 
\end{itemize}
Consequently, existing paradigms struggle to simultaneously achieve modality-invariant representation learning and fine-grained discrimination under the emerging unaligned CM-FGOR setting.

To address these limitations and the lack of benchmarks for unaligned CM-FGOR, we construct a \textbf{F}ine-\textbf{G}rained \textbf{O}ptical-\textbf{S}AR \textbf{a}ircraft and \textbf{s}hip dataset, termed \textbf{FGOS-as}, covering both aerospace and maritime objects. FGOS-as introduces realistic challenges, including large object-scale variation, complex background clutter, severe spatial misalignment, and the modality discrepancy between optical and SAR imagery. To tackle this task, motivated by the inherent discrete-continuous gap between the two modalities, we propose GeoMamba, a geometry-driven, dual-stream retrieval framework. As depicted in Fig. \ref{fig:intro_comparison}(b), GeoMamba combines explicit geometric guidance with the efficient global modeling capability of State Space Models (SSMs). 
Specifically, the dual-stream MambaVision backbone captures long-range spatial dependencies to robustly handle unaligned instances without requiring strict registration. To bridge the profound discrete-continuous gap, a Geometric Feature Injection (GFI) module incorporates structural priors exclusively into the SAR representations, mitigating speckle noise before promoting bidirectional cross-modal semantic interaction. Furthermore, a Geometric Consistency Constraint (GCC) coupled with a Deep Supervision (DS) strategy imposes hierarchical geometric constraints using modality-specific classical operators, \emph{i.e.}, the Sobel operator for optical contours and the Harris detector for SAR scattering centers. This design explicitly aligns modality-invariant geometric structures, enabling the model to focus on intrinsic object structures rather than background clutter and speckle noise.



Extensive experiments show that incorporating explicit structural priors is beneficial for reducing cross-modal discrepancy and suppressing spurious associations between optical clutter and SAR speckle responses. As a result, the proposed GeoMamba achieves superior retrieval performance on the proposed benchmark and demonstrates strong robustness in challenging optical-SAR fine-grained retrieval scenarios.


In summary, our main contributions are as follows:
\begin{enumerate}
    \item[1)] We construct a new fine-grained optical-SAR dataset, termed FGOS-as for cross-modal fine-grained object retrieval (CM-FGOR). The dataset contains 65,646 images across 11 fine-grained aircraft and ship categories, filling the gap in unaligned optical-SAR retrieval.
    \item[2)] We propose GeoMamba, a fine-grained cross-modal retrieval framework built upon State Space Models (SSMs), which efficiently captures long-range spatial dependencies to robustly overcome severe spatial misalignment, thereby enabling effective cross-modal representation learning.
    \item[3)] We design a Geometric Feature Injection (GFI) module to incorporate structural priors, effectively bridging the inherent discrete-continuous modality gap. In addition, a Geometric Consistency Constraint (GCC) module, coupled with a deep supervision strategy, is introduced to distill modality-invariant geometric signatures using explicit edge and corner operators.
\end{enumerate}

In the rest of the paper, we present the related work in Section~\ref{sec:relatedwork} and the proposed FGOS-as dataset in Section~\ref{sec:dataset}. In Section~\ref{sec:methodology}, we elaborate on the geometry-driven GeoMamba framework. Then, experimental settings and results are presented in Section~\ref{sec:Experiments}. Finally, conclusions are drawn in Section~\ref{sec:Conclusion}.

\section{Related Work}
\label{sec:relatedwork}

This section reviews three lines of research closely related to this work: cross-modal image retrieval, fine-grained object recognition, and geometry-aware representation learning.

\subsection{Cross-modal Image Retrieval}
Cross-modal image retrieval aims to match observations acquired from heterogeneous sensors, such as optical and SAR imagery. Existing methods can be broadly categorized into traditional machine learning methods and deep learning methods, the latter mainly including CNN-based, Transformer-based, and State Space Model (SSM)-based frameworks.

\subsubsection{Traditional Machine Learning Methods} 
Early studies mainly relied on subspace learning and metric learning to project heterogeneous features into a common latent space, such as CCA~\cite{CCA} and KCCA~\cite{KCCA}. Subsequent methods, including LMNN~\cite{LMNN}, CMSSH~\cite{CMSSH}, CVH~\cite{CVH}, and GMA~\cite{GMA}, further improved retrieval performance by incorporating semantic priors, hashing strategies, or subspace clustering. However, these methods largely depend on handcrafted descriptors such as SIFT~\cite{SIFT} and LBP~\cite{LBP}, which are often sensitive to viewpoint changes, background clutter, and speckle noise in SAR imagery.

\subsubsection{CNN and Transformer-based Methods} 
Deep learning has advanced cross-modal retrieval by enabling end-to-end feature learning. CNN-based methods commonly adopt two-stream architectures to learn modality-specific features and shared embedding spaces. Representative studies have explored modality balancing~\cite{liu2020parameter}, multi-level feature alignment~\cite{liang2024bridging}, dense correspondence learning~\cite{park2021learning}, and attention mechanisms~\cite{ye2021deep}. Despite their effectiveness, CNN-based models may be limited in efficiently modeling the long-range spatial dependencies required for unaligned remote sensing imagery. 
Transformer architectures, represented by ViT~\cite{dosovitskiy2020image} and DeiT~\cite{touvron2021training}, provide global context modeling through self-attention and have inspired many retrieval models for heterogeneous remote sensing data. Representative studies have explored side information embedding~\cite{he2021transreid}, multi-scale feature variations~\cite{zheng2024versatile}, semantic-controlled representation learning~\cite{chen2023beyond}, and sensor-specific bias reduction~\cite{wang2025cross}. However, Transformer-based models struggle to maintain fine-grained geometric details and the tokenization process tends to blur subtle morphological cues, particularly in fine-grained object patches. Therefore, balancing global semantic modeling and fine-grained structural preservation remains a key challenge in cross-modal retrieval.

\subsubsection{State Space Model-based Methods} 
State Space Models (SSMs) have recently emerged as an efficient architecture for long-range visual modeling. By connecting continuous-time dynamics with discrete sequence processing through discretization, modern SSMs provide an alternative to convolutional and attention-based architectures for modeling long-range dependencies~\cite{gu2021efficiently}. Extensions from 1D sequences to 2D visual data have been explored by multidimensional variants such as S4ND~\cite{nguyen2022s4nd}. More recently, Vision Mamba (ViM)~\cite{zhu2024vision} and VMamba~\cite{liu2024vmamba} introduced spatial scanning mechanisms for image understanding, while MambaVision~\cite{2025mambavision} adopted hierarchical designs for visual representation learning. SSMs have shown promising results in image classification~\cite{zhu2024vision,liu2024vmamba,MambaHSI} and segmentation~\cite{zhu2024samba,chen2024rsmamba}, yet their potential for cross-modal remote sensing retrieval remains largely unexplored. Their ability to capture long-range dependencies with favorable computational efficiency makes them promising for optical-SAR retrieval, where both global contextual understanding and fine-grained structural discrimination are essential.

\subsection{Fine-grained Object Recognition}
Beyond coarse category-level retrieval, fine-grained object recognition requires distinguishing subtle differences among visually similar targets. Early handcrafted descriptors were limited in capturing discriminative local textures and structural cues~\cite{ye}. Deep learning methods have improved fine-grained recognition, yet preserving subtle local details while modeling global context remains challenging. In particular, Transformer-based models~\cite{dosovitskiy2020image} provide strong global representations but may lose fine-grained cues when token abstraction is aggressive. Instance-level retrieval tasks such as person and vehicle re-identification provide useful insights for fine-grained discrimination under domain variations. For example, TransReID~\cite{he2021transreid} reduces sensor and camera bias, VersReID~\cite{zheng2024versatile} improves robustness to multi-scale variations, and SOLIDER~\cite{chen2023beyond} addresses domain gaps through semantic-controlled representation learning. More recently, HOSS-ReID~\cite{wang2025cross} investigates hybrid-sensor ship retrieval in remote sensing scenarios. These studies indicate the importance of preserving identity-related structural cues. Nevertheless, fine-grained optical-SAR retrieval remains challenging due to spatial misalignment, modality-specific noise, and subtle inter-class structural differences.

\subsection{Geometry-aware Representation Learning}
Multi-source remote sensing imagery captures distinct spatial and geometric signatures of observed scenes, which often differ from the appearance statistics of natural images~\cite{ye2017robust}. Consequently, purely data-driven models may provide limited structural interpretability and often fail to fully exploit modality-specific spatial priors, such as continuous optical contours or discrete SAR scattering centers. To address this, geometry-aware and structurally-guided learning have recently attracted increasing attention in remote sensing~\cite{wang2023new}. In object recognition and scene understanding, existing studies have incorporated domain-specific priors, such as spatial topology~\cite{yi2024deep}, structural keypoints~\cite{xiong2025sar}, and geometric constraints~\cite{yang2025deep}, into deep networks to guide feature learning. These studies demonstrate that integrating explicit structural knowledge can improve model robustness and bridge the gap between data-driven learning and inherent sensing characteristics. 
However, many of these prior-guided methods still rely on explicit modeling pipelines or additional preprocessing, which hinders efficient end-to-end optimization and deployment. In contrast, the use of lightweight geometric priors for unpaired optical-SAR fine-grained retrieval remains largely underexplored.

\section{FGOS-as: An Optical-SAR Fine-grained Image Retrieval Dataset}
\label{sec:dataset}

\begin{figure*}[t]
    \centering
        \resizebox{1\textwidth}{!}{
        \begin{tabular}{cc}
            \multicolumn{2}{c}{\includegraphics[width=1\textwidth, valign=m]{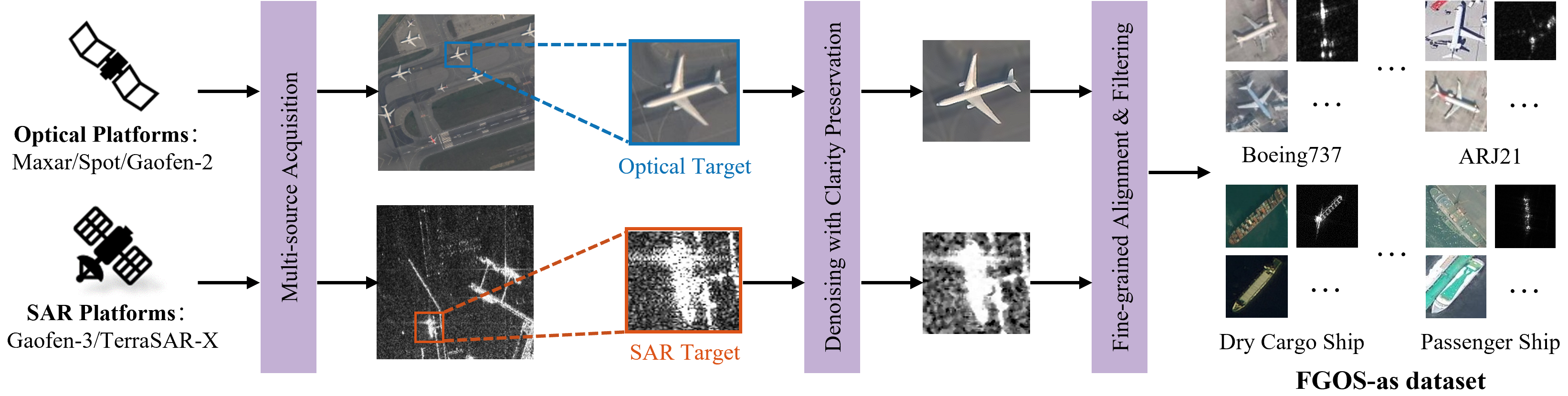}} \\
             \multicolumn{2}{c}{\small{(a) Construction workflow of the FGOS-as dataset}}  \\          
            \noalign{\smallskip}
            
            \includegraphics[width=0.66\textwidth, valign=m]{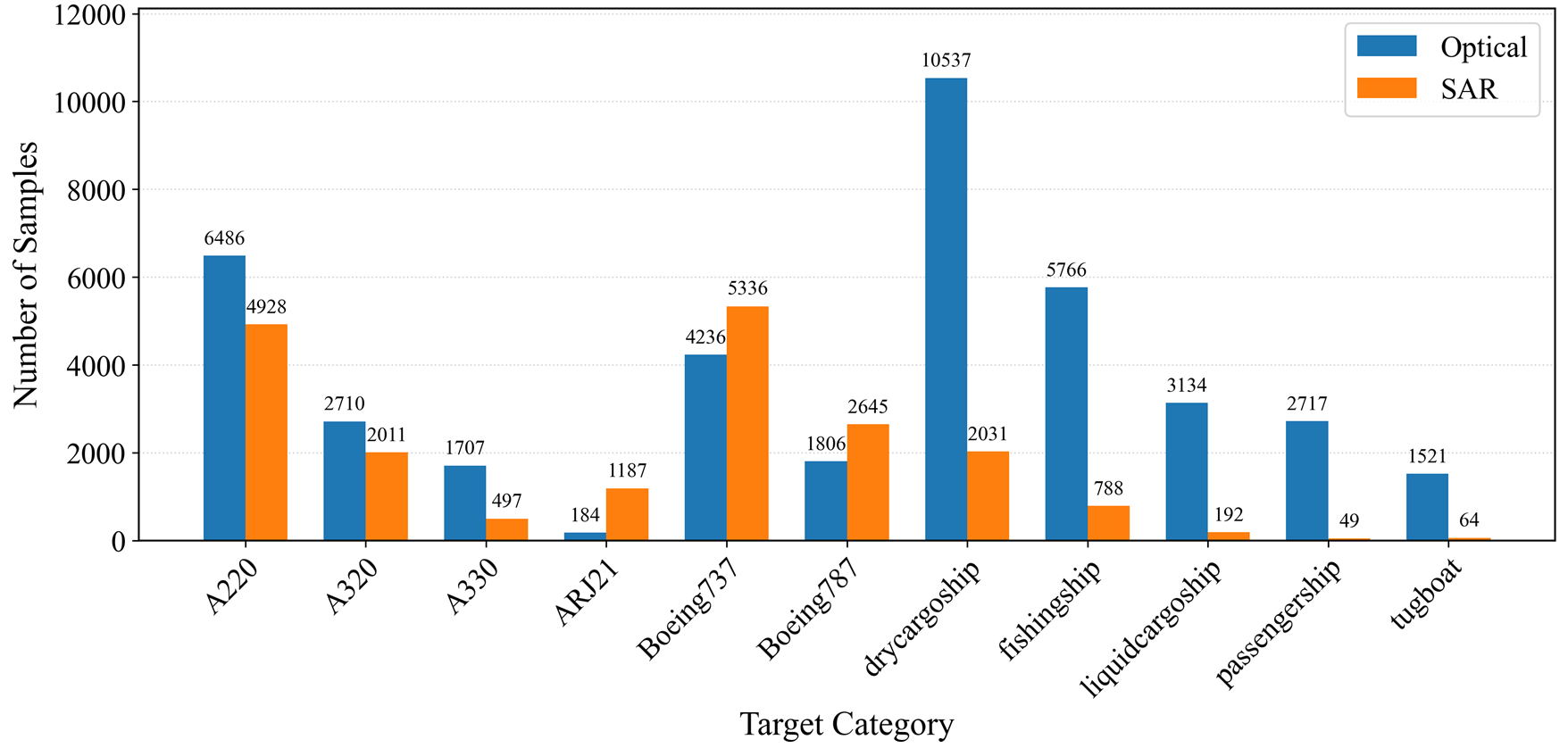} &
            \includegraphics[width=0.33\textwidth, valign=m]{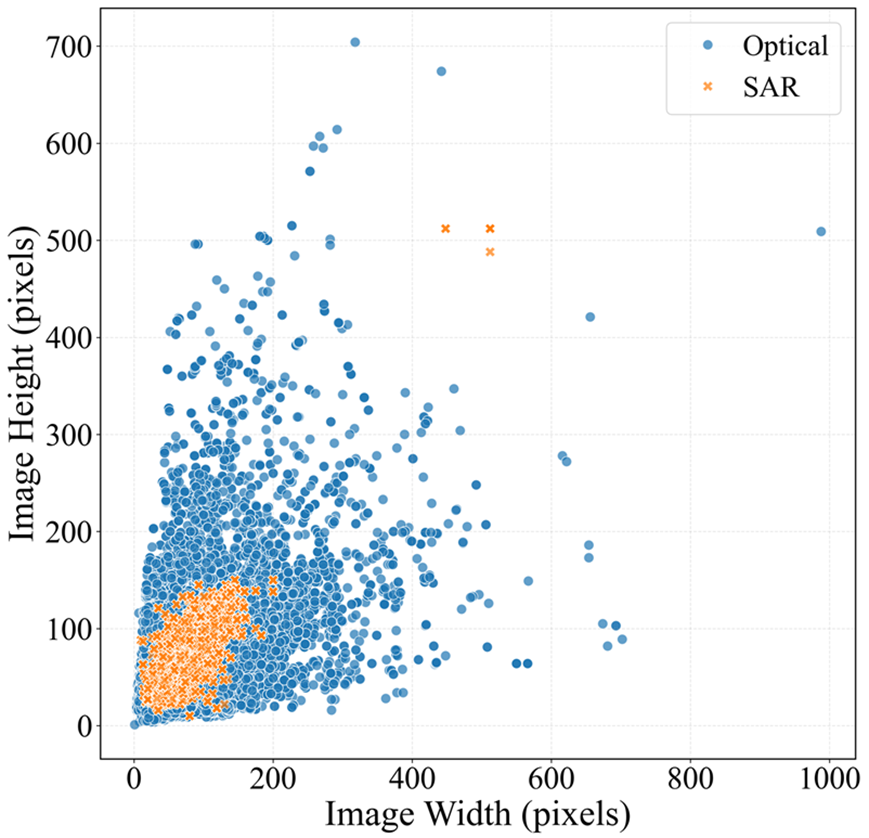} \\
            \small{(b) Category distribution of the dataset}&\small{(c) Resolution distribution of the dataset}\\
        \end{tabular}
    }
    \caption{Overview of the proposed FGOS-as dataset. (a) Workflow of the FGOS-as dataset construction, illustrating the stages of multi-source acquisition, denoising with clarity preservation, and fine-grained classification. (b) Category distribution of the clean target subset, excluding distractor samples, detailing the varying number of instances across the 11 fine-grained aerospace and maritime categories. (c) Resolution distribution, visualizing the diverse image dimensions for both optical and SAR modalities.}
    \label{fig:dataset_overview}
\end{figure*}

To advance practical cross-modal object retrieval, we construct a new dataset termed \textbf{FGOS-as} (\textbf{F}ine-\textbf{G}rained \textbf{O}ptical-\textbf{S}AR \textbf{a}ircraft and \textbf{s}hip dataset). It is specifically designed for the fine-grained identification of aerospace and maritime targets across optical and SAR modalities. Unlike conventional re-identification datasets~\cite{8510891,di2021public,yuming2025osdataset2} that assume homogeneous data or rely on perfectly aligned image pairs, FGOS-as pioneers fine-grained object retrieval under realistic, spatially and temporally unaligned optical-SAR conditions and introduces cross-modal discrepancies. As illustrated in Fig.~\ref{fig:dataset_overview}(a), FGOS-as is constructed through a three-stage pipeline tailored to the practical sensing characteristics of heterogeneous imagery, \emph{i.e.}, multi-source acquisition, denoising with clarity preservation, and fine-grained alignment \& filtering.

\begin{table*}[htbp]
    \centering
    \caption{Comparisons of Several Publicly Available Remote Sensing Image Retrieval Datasets}
    \label{tab:datasets}
   \setlength{\tabcolsep}{10pt}
 
    \resizebox{\textwidth}{!}{%
    \renewcommand{\arraystretch}{1.2} 
    \begin{tabular}{l c c c c l c} 
        \toprule
        Dataset & Year & Modalities & Granularity & Alignment & Scenario & Resolution \\
        \midrule
        
        UCMD~\cite{yang2010bag} & 2010 & RGB & Coarse & -- & Scene Classification & 0.3m \\
        
        WHU-RS19~\cite{xia2010structural} & 2010 & RGB & Coarse & -- & Land Use Analysis & Up to 0.5m \\
       
        DSRSID~\cite{li2018learning} & 2018 & PAN, RGB, NIR & Coarse & Spectral & CM Retrieval & 2m / 8m \\
        
        CBRSIR\_VS~\cite{sun2021deep} & 2021 & RGB, SAR & Coarse & Coarse & Cross-modal Retrieval & 1m / 10m \\

        MRSSID~\cite{xiong2022interpretable} & 2022 & PAN, MSI & \textbf{Fine} & Spectral & Cross-modal Retrieval & 1m / 4m \\
         
        FGSCR42~\cite{di2021public} & 2021 & RGB & \textbf{Fine} & -- & Object Retrieval & N/A \\
        
        SOPatch~\cite{xu2023sar} & 2023 & RGB, SAR & N/A & Patch-level & Feature Matching & 1m $\sim$ 10m \\
        
        OSDataset2.0~\cite{yuming2025osdataset2} & 2025 & RGB, SAR & N/A & Pixel-level & Image Matching & 0.43m $\sim$ 1m \\

        M4-SAR~\cite{wang2025m4} & 2025 & RGB, SAR & Coarse & Instance-level & Object Detection  & 10m, 60m  \\
        
        \midrule
        
        \textbf{FGOS-as (Ours)} & 2026 & RGB, SAR & \textbf{Fine} & \textbf{Unaligned} & \textbf{CM-FGOR} & 0.5m $\sim$ 3.0m  \\
        
        \bottomrule
    \end{tabular}%
    }
\end{table*}

\subsection{Comparative Analysis of Datasets}
Table~\ref{tab:datasets} summarizes representative remote sensing retrieval benchmarks. These datasets have substantially advanced research in scene retrieval, object retrieval, and cross-modal matching, yet they remain inadequate for studying spatially and temporally unaligned CM-FGOR. Widely used benchmarks such as UCMD~\cite{yang2010bag} and WHU-RS19~\cite{xia2010structural} mainly focus on scene classification or coarse category retrieval, lacking object-centric annotations for distinguishing visually similar objects. Fine-grained datasets, such as FGSCR42~\cite{di2021public} and MRSSID~\cite{xiong2022interpretable}, improve annotation granularity but are typically limited to single-modal imagery or a single target domain. In the area of cross-modal matching, datasets like DSRSID~\cite{li2018learning} and CBRSIR\_VS~\cite{sun2021deep} laid the foundation, while more recent contributions like SOPatch~\cite{xu2023sar} and OSDataset2.0~\cite{yuming2025osdataset2} have significantly promoted local feature matching and pixel-level image registration. However, these datasets, together with detection-oriented benchmarks like M4-SAR~\cite{wang2025m4}, are constructed from spatially aligned or pairwise matched samples, where cross-modal correspondence is pre-established through strict spatial registration. Such settings do not fully reflect modern multi-constellation sensing scenarios, in which optical and SAR images are often acquired at different times, viewing angles, and spatial resolutions, resulting in temporal inconsistency and geometric misalignment.

To address these limitations, we construct FGOS-as, a benchmark for unaligned CM-FGOR. It contains 65,646 optical and SAR images covering both aerospace and maritime objects, with fine-grained annotations across 11 categories and substantial variations in scale, viewpoint, and acquisition conditions. By explicitly incorporating realistic spatiotemporal discrepancies and modality gaps, FGOS-as provides a comprehensive benchmark for developing robust CM-FGOR methods.

\begin{table}[t]
\centering
\caption{Data Sources and Sensor Specifications of FGOS-as.}
\label{tab:data_sources}
\small
\renewcommand{\arraystretch}{1.2}
\setlength{\tabcolsep}{4pt}
\resizebox{\columnwidth}{!}{
\begin{tabular}{c c c c}
\toprule
\textbf{Modality} & \textbf{Source Dataset} & \textbf{Sensor/Platform} & \textbf{Band} \\
\midrule
\multirow{2}{*}{Optical} & Google Earth & Maxar/Spot & RGB \\
 & FAIR1M 2.0~\cite{sun2022fair1m} & Gaofen-2/WorldView & RGB \\
\midrule
\multirow{4}{*}{SAR} & FUSAR-Ship \cite{hou2020fusar} & Gaofen-3 & C-band \\
 & Sar-aircraft-1.0~\cite{zhirui2023sar} & Gaofen-3 & C-band \\
 & Air-sarship~\cite{xian2019air} & Gaofen-3 & C-band \\
 & FAIR-CSAR~\cite{10806766} & TerraSAR-X & X-band \\
\bottomrule
\end{tabular}
}
\end{table}

\subsection{Dataset Construction}
\subsubsection{Multi-source Acquisition}
To ensure diversity in sensing conditions and object appearance, we collect raw data from multiple public sources rather than a single platform.  
As detailed in Table~\ref{tab:data_sources}, optical imagery is obtained from Google Earth and FAIR1M~\cite{sun2022fair1m}, providing rich texture and illumination diversity. SAR imagery is collected from multi-band constellations, including Gaofen-3 and TerraSAR-X, with samples drawn from FUSAR-Ship~\cite{hou2020fusar}, Sar-aircraft-1.0~\cite{zhirui2023sar}, Air-sarship-1.0~\cite{xian2019air}, and FAIR-CSAR~\cite{10806766}. 
This multi-source acquisition strategy introduces substantial variation in spatial resolution (0.5m to 3.0m), incidence angle, and imaging conditions, thereby creating realistic challenges for CM-FGOR.

\subsubsection{Denoising with Clarity Preservation}
As illustrated in the workflow, object instances (i.e., Optical and SAR targets) are first cropped from the raw multi-source scenes according to bounding boxes from the original annotation files. Following this target extraction, we implement modality-specific pre-processing on these cropped patches to improve visual quality while preserving intrinsic object characteristics.
\begin{itemize}
    \item For optical imagery, we employ an edge-preserving bilateral filter~\cite{chen2020gaussian} to suppress local clutter while retaining object boundaries, followed by Laplacian sharpening~\cite{ma2014optimized} to moderately enhance structural contours.
    \item For SAR imagery, we apply a median filter~\cite{qiu2004speckle} for basic despeckling, followed by contrast limited adaptive histogram equalization (CLAHE)~\cite{mohammed2025contrast} to highlight the discrete scattering centers. A simple thresholding step is further used to suppress weak background responses while preserving salient structural structures.
\end{itemize}
These operations are uniformly applied to the specific modality across the dataset and are intended only to improve sample readability without altering semantic labels.

\subsubsection{Fine-grained Alignment \& Filtering}
After the denoising stage, all enhanced samples undergo modality-specific manual quality control. Specifically, we discard optical images with severe occlusion (\textgreater 50\%) or poor clarity, and SAR patches with extreme noise or ambiguous target structure. Subsequently, each remaining object is assigned a fine-grained label (e.g., Boeing737, Dry Cargo Ship) based on source metadata to construct the unified FGOS-as dataset.

Finally, object-centered cropping is independently applied to each sample to place the object in the center with a proportional margin, while explicitly preserving its native spatial resolution and realistic scale variations. Note that no pixel-level registration or pairwise geometric alignment is performed between optical and SAR samples, thus retrieval setting remains spatially and temporally unaligned.

\subsection{Analysis of Dataset Properties}

\subsubsection{Category and Resolution Distribution}
The final FGOS-as retrieval benchmark contains 65,646 images, including clean fine-grained target samples and additional distractor samples introduced to simulate realistic large-scale retrieval scenarios. Among them, the clean target subset covers 11 fine-grained aerospace and maritime categories, whose category-wise distribution is summarized in Fig.~\ref{fig:dataset_overview}(b). Moreover, as illustrated in Fig.~\ref{fig:dataset_overview}(c), 
rather than relying on uniformly cropped patches with fixed-size inputs, 
FGOS-as includes samples with varying widths, heights, and ground sampling resolutions across both optical and SAR modalities. Such variability better reflects real-world remote sensing acquisition conditions, where object appearance often changes significantly across sensors and scenes. 

\subsubsection{Intrinsic Challenges}
In general, compared with existing datasets, FGOS-as presents three key challenges:
\begin{itemize}
    \item \textbf{Discrete-continuous modality gap}: Optical imagery describes continuous surface appearance, whereas SAR imagery is dominated by discrete geometric keypoints. This fundamental difference in imaging mechanisms makes direct cross-modal feature matching highly challenging.
    \item \textbf{Complex background clutter}: The dataset includes challenging scenes such as airport aprons, harbors, and container terminals, where dense surrounding structures and rich background textures can distract object localization and degrade reliable cross-modal correspondence.
    \item \textbf{Large multi-scale variation}: The dataset contains targets ranging from large aircraft (\emph{e.g.}, A330) to compact vessels (\emph{e.g.}, fishing boats), leading to substantial scale diversity that requires hierarchical perception and scale-adaptive representation learning.
\end{itemize}

These characteristics make FGOS-as a challenging and realistic benchmark for evaluating retrieval models under heterogeneous sensing conditions, large modality discrepancies, and real-world acquisition misalignment.

\begin{figure*}[htbp]
    \centering
    \includegraphics[width=1\textwidth]{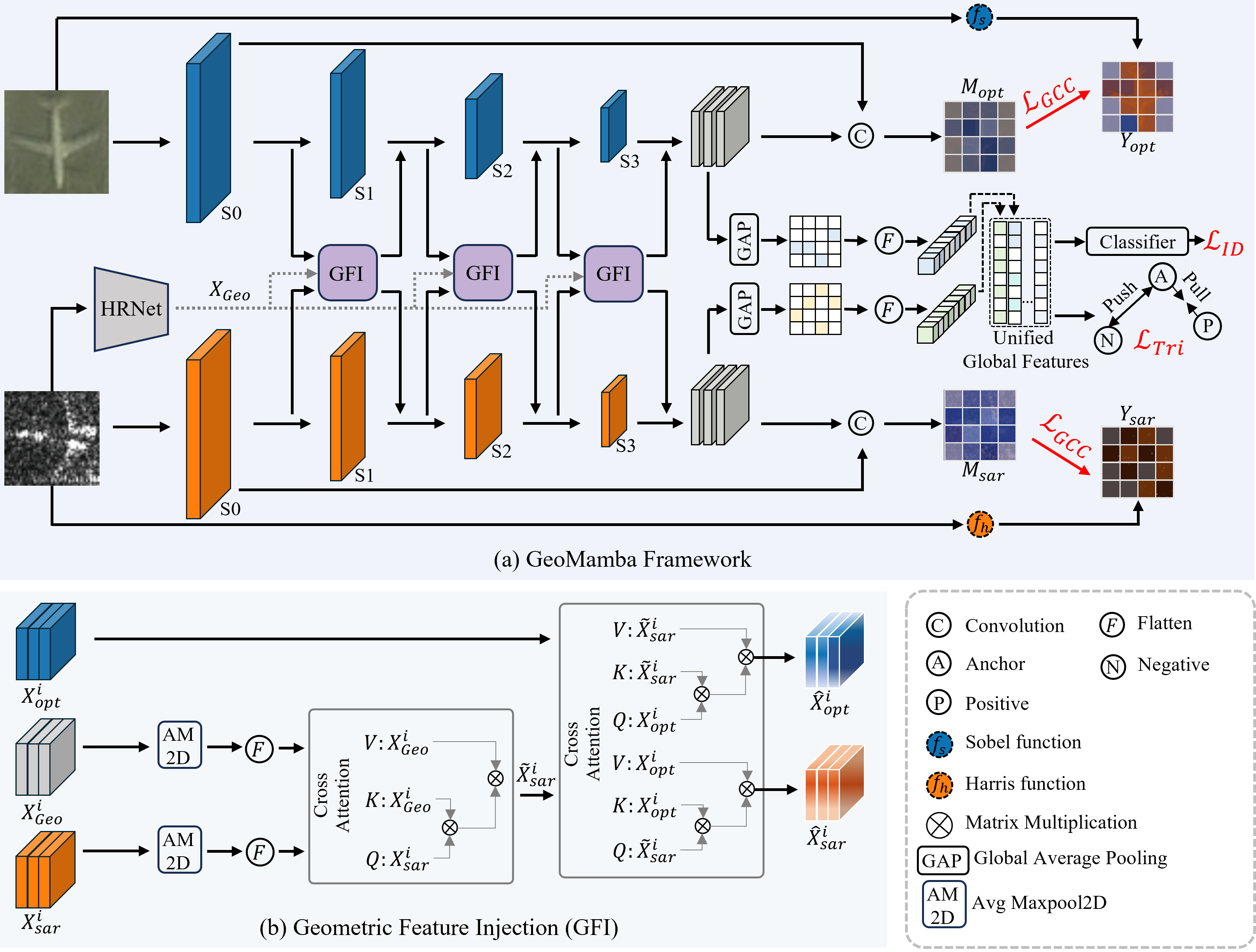}
    \caption{Architecture of the proposed method. (a) GeoMamba framework, which features a dual-stream MambaVision backbone for extracting spatial representations from optical and SAR inputs alongside a GCC constraint enforcing geometric consistency, and incorporates the (b) GFI module at intermediate stages to bridge the modality gap through structural prior injection integrating auxiliary HRNet geometric priors with SAR representations to mitigate speckle noise, followed by cross-modal semantic interaction facilitating bidirectional information exchange between the enhanced SAR and optical features.}
    \label{fig:overall_architecture}
\end{figure*}

\section{Proposed Method}
\label{sec:methodology}

In this section, we present the proposed GeoMamba for cross-modal fine-grained optical-SAR object retrieval. We first formulate the CM-FGOR task and then provide an overview of our framework. Next, we dissect the individual components and the associated objective functions for model optimization.

\subsection{Formulation of CM-FGOR}

In this work, we focus on optical-SAR retrieval, where images are acquired by different sensors under varying acquisition times, imaging geometries, and resolutions. Therefore, the two modalities are not assumed to be spatially registered or temporally paired. Specifically, let the optical and SAR datasets be denoted as 
$\mathcal{D}^{opt} = \{(x_i^{opt},y_i)\}_{i=1}^{N_{opt}}$ and $\mathcal{D}^{sar} =\{(x_j^{sar},y_j)\}_{j=1}^{N_{sar}}$, respectively, where
$x_i^{opt}$ and $x_j^{sar}$ are object-centered image patches, and
$y$ denotes the fine-grained identity label shared across modalities. 
Given a query image from one modality, the objective of CM-FGOR is to retrieve gallery images containing the mixed modality data that belong to the same label.

Unlike conventional retrieval paradigms that operate within a single modality, CM-FGOR aims to learn feature representations that are modality-invariant and fine-grained discriminative under unpaired and unaligned optical-SAR conditions. To optimize the retrieval model, our objective is to map heterogeneous samples into a shared embedding space, where samples belonging to the same fine-grained category are pulled closer regardless of modality, while samples from different categories are pushed farther apart. Similarity between two embeddings is measured by the Euclidean distance $d(f_i, f_j) = \|f_i - f_j\|_2$.

Formally, for an input sample $x_i$ in a mini-batch of size $N$, let $f_i = \Phi(x_i; w)$ denote its feature extracted by the model $\Phi$ parameterized by $w$. Given the identity labels $\{y_i\}_{i=1}^N$, the overall retrieval objective is defined as:
\begin{equation}
\label{equ:retrieval}
    \mathcal{L}_{Retrieval} = \frac{1}{N} \sum_{i=1}^{N} \mathcal{L}_{id}(\phi(f_i), y_i) + \frac{\lambda_{tri}}{N} \sum_{i=1}^{N} \mathcal{L}_{tri}(f_i),
\end{equation}
where $\mathcal{L}_{id}$ is the identity loss, $\mathcal{L}_{tri}$ is the triplet loss, $\phi$ is the prediction head and $\lambda_{tri}$ is the balancing weight. The identity loss enhances category-level discrimination, while the triplet loss reduces cross-modal intra-class discrepancy and enlarges inter-class separation. 
For each anchor feature $f_i$, the triplet loss is formulated as:
\begin{equation}
\begin{aligned}    
\mathcal{L}_{tri}(f_i) &= \max \left(0, d_{i,p} - d_{i,n} + m \right),\\
d_{i,p} &= \max_{j: y_j = y_i} d(f_i, f_j), \\
d_{i,n} &= \min_{j: y_j \neq y_i} d(f_i, f_j),
\end{aligned}
\end{equation}
where $m$ is the margin, $d_{i,p}$ and $d_{i,n}$ denote the distances to the hardest positive and hardest negative samples, respectively. 
By jointly optimizing these objectives, the model learns a unified feature space that preserves fine-grained target identity.


\subsection{Framework Overview}
To overcome optical-SAR retrieval without spatial registration, we propose GeoMamba (see Fig.~\ref{fig:overall_architecture}), consisting of three components: (1) A dual-stream MambaVision backbone that captures long-range global dependencies in unaligned optical-SAR data; (2) a Geometric Feature Injection (GFI) module that bridges the modality gap by enhancing modality-invariant geometric structures; and (3) a Geometric Consistency Constraint (GCC) coupled with Deep Supervision (DS) to learn hierarchical representations to suppress background clutter and handle large-scale variations.

Specifically, given unaligned optical and SAR inputs $x_{opt}$ and $x_{sar}$, the dual-stream backbone extracts intermediate features $X_{opt}^{in}$ and $X_{sar}^{in}$ in parallel. To reduce the discrepancy between continuous optical appearances and discrete SAR geometric responses, GFI introduces structural priors $X_{Geo}^{in}$ extracted by an auxiliary HRNet exclusively into the SAR branch. Unlike optical images, this asymmetric injection mitigates inherent SAR speckle noise and regularizes discrete structural configurations. Building upon this enhanced SAR representation ($\tilde{X}_{sar}$), GFI then performs cross-modal feature interaction with $X_{opt}^{in}$. 
Furthermore, to filter background clutter and retain object shapes, DS uses a convolutional head to map shallow and deep features into single-channel spatial masks, retaining fine-grained geometric structures before global pooling.

The overall network is trained using a joint objective composed of identity loss, triplet loss, and the proposed GCC loss. Guided by modality-specific geometric signals, \emph{i.e.}, Sobel operator for continuous optical contours from $x_{opt}$ and Harris detector for discrete geometric keypoints from $x_{sar}$, GCC encourages the model to focus on informative object structures rather than background clutter, thereby learning discriminative, modality-invariant representations for CM-FGOR.

\subsection{Dual-stream MambaVision Backbone}
To process the inputs $x_{opt}$ and $x_{sar}$, two separated modality-specific patch embedding layers first project them into shallow feature maps, preserving sensing-specific low-level characteristics. The two streams then pass through hierarchical MambaVision blocks to extract multi-level representations, accommodating the challenge of large multi-scale variations. Across different stages, convolutional operations are used to capture local textures and fine structural details, while State Space Models (SSMs) based blocks efficiently model long-range spatial dependencies. This hybrid design helps jointly encode local geometric cues and global scene structures, which is particularly crucial for robustly handling unaligned optical and SAR data without relying on strict spatial registration.

During feature extraction, the proposed GFI module is inserted at intermediate stages to specifically bridge the discrete-continuous modality gap. Through progressive downsampling and hierarchical encoding, the backbone produces deep spatial features, simply denoted as $X_{opt}^{deep}$ and $X_{sar}^{deep}$. Before global pooling, these features are explicitly constrained by the DS strategy to retain spatial configurations and filter background clutter. Finally, they are aggregated via global average pooling into semantic embeddings $f_{opt}$ and $f_{sar} \in \mathbb{R}^{1024}$ for the final cross-modal retrieval.

\subsection{Geometric Feature Injection}

In optical-SAR retrieval, the discrete-continuous modality gap and structural degradation limit the effectiveness of independently learned features. To address this issue, we propose the GFI module, which combines structural prior injection for SAR enhancement and subsequent cross-modal semantic interaction for bidirectional information exchange, as detailed in Fig.~\ref{fig:overall_architecture} (b).

\subsubsection{Structural Prior Injection}
Given the extracted feature maps $X_{opt}^i$ and $X_{sar}^i$ at stage $i$, GFI first introduces structural priors to refine the SAR representations. Specifically, an auxiliary HRNet branch~\cite{huang2025physics} is used to extract clean geometric priors $X_{Geo}^i$. This asymmetric injection is structurally motivated: optical images inherently possess continuous surfaces and clear boundaries, enabling the backbone to sufficiently encode their geometric structures without requiring auxiliary enhancement. In contrast, SAR images suffer from severe speckle noise and structural degradation, making it difficult for the backbone to independently capture stable spatial configurations. Therefore, the structural prior exclusively targets the SAR branch to act as a geometric skeleton, mitigating speckle noise and regularizing discrete structural configurations. To integrate these priors, we design an asymmetric fusion block based on cross-attention. The pooled SAR feature serves as the query ($Q$), while the geometric prior provides the key and value ($K, V$):
\begin{equation}
    \tilde{X}_{sar}^i = X_{sar}^i + \text{CrossAttn}(X_{sar}^i, X_{Geo}^i),
\end{equation}
where $\tilde{X}_{sar}^i$ denotes the geometry-enhanced SAR feature. This design allows the SAR branch to selectively exploit informative structural cues, thereby enhancing robustness against noise.

\subsubsection{Cross-modal Semantic Interaction}
Building upon this geometric enhancement, GFI then performs bidirectional feature interaction to reduce the modality discrepancy and prevent noisy SAR artifacts from interfering with the clear optical features during the semantic exchange. 
To achieve this, the spatial dimensions of $X_{opt}^i$ and the enhanced $\tilde{X}_{sar}^i$ are flattened into token sequences. By alternately treating one modality as the query ($X_q$) and the other as the key/value ($X_{kv}$), Multi-Head Cross-Attention (MHCA)~\cite{vaswani2017attention} is used to aggregate complementary information across modalities. We denote the intermediate representation after the MHCA module and residual connection as $X_q^{\prime}$:
\begin{equation}
    X_q^{\prime} = X_q + \text{MHCA}(X_q, X_{kv}).
\end{equation}
We then define the complete cross-modal interaction mapping, denoted as $\mathcal{F}_{cross}$, to further process this intermediate output via a feed-forward network:
\begin{equation}
    \mathcal{F}_{cross}(X_q, X_{kv}) = X_q^{\prime} + \text{MLP}\Big(\text{LN}\big(X_q^{\prime}\big)\Big),
\end{equation}
where $\text{LN}$ denotes Layer Normalization. Applying this mapping symmetrically, the updated representations are reshaped back to 2D spatial feature maps:
\begin{equation}
\begin{aligned}
    \hat{X}_{opt}^i &= \mathcal{F}_{cross}(X_{opt}^i, \tilde{X}_{sar}^i),\\
    \hat{X}_{sar}^i &= \mathcal{F}_{cross}(\tilde{X}_{sar}^i, X_{opt}^i).
\end{aligned}
\end{equation}
This bidirectional interaction provides cross-modal contextual guidance, allowing both modalities to benefit from semantic exchange while preserving the enhanced structural integrity.

\subsection{Geometric Consistency Constraint}
CM-FGOR requires preserving object structures alongside high-level semantic representations. We introduce a Geometric Consistency Constraint (GCC) using modality-specific pseudo-labels, firmly embedding geometric priors (\emph{e.g.}, edges and geometric corners) into the feature space to resist complex background interference, as illustrated in Fig.~\ref{fig:overall_architecture}.

\subsubsection{Hierarchical Geometric Representation}
Instead of imposing supervision at every stage, GCC is applied only at the shallowest ($i=0$) and deepest ($i=3$) stages. The shallow stage preserves fine-grained structural responses before downsampling blurs object boundaries, while the deepest stage retains explicit geometric configurations prior to global pooling. By constraining only the boundary stages, the framework preserves structural consistency while allowing intermediate representations to adaptively bridge the modality gap.

Specifically, a modality-specific projection head, implemented using a single convolutional layer $\mathcal{H}_{DS}$, projects the stage features into single-channel spatial masks. Given the shallow-stage features $X{opt}^0$ and $X_{sar}^0$, and final-stage deep features denoted as ${X}_{opt}^3$ and ${X}_{sar}^3$, the corresponding auxiliary masks are predicted as:
\begin{equation}
\begin{aligned}
    M_{\{opt, sar\}}^{shallow} &= \mathcal{H}_{DS} ( X_{\{opt, sar\}}^0 ),\\
    M_{\{opt, sar\}}^{deep} &= \mathcal{H}_{DS} ( {X}_{\{opt, sar\}}^3 ),
\end{aligned}
\end{equation}
where $M^{shallow}$ act as an early geometric anchor to capture pristine geometric responses, and $M^{deep}$ forces the high-level features to retain fine-grained spatial configurations immediately before global pooling. This hierarchical mechanism guarantees geometric awareness in the retrieval embeddings.

\subsubsection{Deep Supervision Strategy}
To supervise the predicted masks, we propose a Deep Supervision (DS) strategy to suppress complex background interference and alleviate the discrete-continuous modality gap. Owing to the distinct imaging mechanisms of optical and SAR modalities, modality-specific pseudo-labels are constructed. Optical imagery is characterized by continuous intensity variations, making the Sobel operator effective for extracting structural contours. In contrast, SAR imagery is heavily affected by speckle noise, while man-made structures are typically represented as sparse and discrete scattering centers. Accordingly, the Harris detector is employed to capture salient geometric keypoints in SAR images. The resulting modality-specific pseudo-label masks, serving as geometric supervision targets, are formulated as:
\begin{equation}
\begin{aligned}
    Y_{opt} &= \text{Sobel} ( x_{opt} ), \\
    Y_{sar} &= \text{Harris} ( x_{sar} ),
\end{aligned}
\end{equation}
where $Y_{opt}, Y_{sar} \in \{0, 1\}^{H \times W}$ denote the binary ground-truth masks, $H$ and $W$ are the height and width, respectively.

Since geometric responses typically occupy only a small fraction of the spatial regions, we employ the Focal Loss to mitigate the severe class imbalance between foreground structures and background areas. Accordingly, the hierarchical GCC loss is formulated as:
\begin{equation}
\begin{aligned}
    \mathcal{L}_{GCC} &= \sum_{m \in \{opt, sar\}} \Big( \lambda_{deep} \mathcal{L}_{focal} ( M_m^{deep}, Y_m ) \\
    &\quad + \lambda_{shallow} \mathcal{L}_{focal} ( M_m^{shallow}, Y_m ) \Big),
\end{aligned}
\end{equation}
where $\mathcal{L}_{focal}$ represents the Focal Loss, and $Y_m$ is downsampled to match the spatial resolution of the respective predicted masks. $\lambda_{deep}$ and $\lambda_{shallow}$ control the contributions of deep and shallow supervision. The final training objective is:
\begin{equation}
    \mathcal{L}_{total} = \mathcal{L}_{Retrieval} + \lambda_{GCC} \mathcal{L}_{GCC},
\end{equation}
where $\mathcal{L}_{Retrieval}$ denotes the standard cross-modal retrieval loss defined in Equation~\ref{equ:retrieval}, and $\lambda_{GCC}$ balances the geometric supervision term. This design ensures that the network to learn retrieval representations that preserve structural integrity while effectively resisting modality-specific background interference.

\begin{table*}[t]
\centering
\caption{Quantitative Comparison Of Fine-Grained Cross-Modal Retrieval Performance On FGOS-as Dataset. We Evaluate The mAP (\%) And Rank-$k$ ($k=1, 3, 5$) (\%)  Accuracies Across Different Retrieval Settings. \textbf{Bold} And \underline{Underlined} Values Mark The Best And Second-Best Performances In Each Column.}
\label{tab:comparison}
\small 
\renewcommand{\arraystretch}{1.2}
\setlength{\tabcolsep}{6pt}
\resizebox{\textwidth}{!}{
\begin{tabular}{c cccc cccc cccc}
\toprule
\multicolumn{1}{c}{\multirow{2}{*}[-2pt]{Methods}} & \multicolumn{4}{c}{All-to-All} & \multicolumn{4}{c}{Optical-to-SAR} & \multicolumn{4}{c}{SAR-to-Optical} \\ 
\cmidrule(lr){2-5} \cmidrule(lr){6-9} \cmidrule(lr){10-13} 
\multicolumn{1}{c}{} & \multicolumn{1}{c}{mAP} & \multicolumn{1}{c}{Rank-1} & \multicolumn{1}{c}{Rank-3} & \multicolumn{1}{c}{Rank-5}  
 & \multicolumn{1}{c}{mAP} & \multicolumn{1}{c}{Rank-1} & \multicolumn{1}{c}{Rank-3} & \multicolumn{1}{c}{Rank-5}  
 & \multicolumn{1}{c}{mAP} & \multicolumn{1}{c}{Rank-1} & \multicolumn{1}{c}{Rank-3} & \multicolumn{1}{c}{Rank-5} \\ 
\midrule

SOLIDER    \cite{chen2023beyond}     & 34.6 & 45.9 & 63.9 & 71.2 & 15.2 & 17.7 & 34.2 & 43.9 & 30.9 & 37.7 & 44.9 & 49.6 \\

TransReID  \cite{he2021transreid}    & 38.1 & 47.2 & 68.6 & 74.3 & 31.2 & 37.8 & 56.2 & 65.2 & 34.6 & 40.5 & 47.6 & 51.1 \\

DEEN       \cite{Zhang_2023_CVPR}    & 41.2 & 50.1 & 71.3 & 76.8 & 21.1 & 23.3 & 44.1 & 55.4 & 59.7 & 60.1 & 69.3 & 73.5 \\

VersReID   \cite{zheng2024versatile} & 48.6 & 57.7 & 78.4 & 82.5 & 32.2 & 41.2 & 58.7 & 67.4 & 57.4 & 59.9 & 66.2 & 71.3 \\

D2InterNet \cite{liu2025advancing}   & 44.6 & 52.9 & 74.5 & 79.2 & 32.9 & 43.5 & 61.3 & 69.8 & 55.6 & 58.7 & 64.6 & 69.5 \\

TransOSS   \cite{wang2025cross}      & \underline{58.6} & \underline{72.1} & \underline{81.9} & \underline{85.4} & \underline{37.9} & \underline{54.3} & \underline{64.5} & \underline{71.2} & \underline{60.4} & \underline{60.9} & \underline{70.5} & \underline{74.7} \\

\textbf{GeoMamba (Ours)} & \textbf{63.3} & \textbf{77.0} & \textbf{84.4} & \textbf{87.3} & \textbf{42.1} & \textbf{62.5} & \textbf{69.2} & \textbf{74.8} & \textbf{67.8} & \textbf{64.8} & \textbf{73.4} & \textbf{75.4} \\ 

\rowcolor[gray]{.9} \textit{w.r.t. SOTA} & +4.7 & +4.9 & +2.5 & +1.9 & +4.2 & +8.2 & +4.7 & +3.6 & +7.4 & +3.9 & +2.9 & +0.7 \\
\bottomrule
\end{tabular}
}
\end{table*}

\section{Experiments}
\label{sec:Experiments}

In this section, we evaluate the proposed GeoMamba framework. We first describe the experimental settings on the newly constructed FGOS-as dataset, followed by quantitative and qualitative comparisons with state-of-the-art methods. Finally, we conduct ablation studies to assess the key components.

\subsection{Experimental Settings} 

\subsubsection{Datasets} 
To evaluate the proposed method, we use the newly constructed FGOS-as dataset for cross-modal fine-grained aircraft and ship retrieval. The dataset contains 65,646 images across 11 fine-grained categories, including specific types of aircraft and ships. It is divided into a training set with 39,336 images, a query set with 7,289 images, and a gallery set with 19,021 images. A retrieved sample is considered correct if it belongs to the same fine-grained category as the query. 

\subsubsection{Baseline Methods} 
We compare the proposed GeoMamba with several representative state-of-the-art (SOTA) methods from three categories: \emph{i}) General image retrieval and re-identification methods, including TransReID~\cite{he2021transreid}, VersReID~\cite{zheng2024versatile}, SOLIDER~\cite{chen2023beyond}, and D2InterNet~\cite{liu2025advancing}; \emph{ii}) Cross-modal visible-infrared retrieval methods, including DEEN~\cite{Zhang_2023_CVPR}; and \emph{iii})  Remote sensing optical-SAR retrieval methods, including TransOSS~\cite{wang2025cross}.

\subsubsection{Evaluation Metrics} 
Following standard retrieval protocols, we adopt mean Average Precision (mAP) and Rank-$k$ accuracy as evaluation metrics. Specifically, mAP measures overall retrieval quality across all queries, while Rank-$k$ reports the probability that at least one correct match appears within the top-$k$ retrieved results. We establish three evaluation settings in the experiments, namely All-to-All, Optical-to-SAR, and SAR-to-Optical. Specifically, All-to-All denotes a unified retrieval protocol where any optical or SAR image is used as a query against a gallery containing both optical and SAR images. Optical-to-SAR uses optical images as queries to retrieve semantically corresponding fine-grained targets from a SAR gallery, while SAR-to-Optical performs the reverse process by using SAR queries to retrieve targets from an optical gallery.

\subsubsection{Implementation Details} 
We employ several data augmentation strategies to prevent overfitting, including random horizontal flipping, padding, and random dropping. All optical and SAR images are resized to $224\times224$ for training. The MambaVision backbone is initialized with weights pre-trained on ImageNet~\cite{deng2009imagenet}. The model is implemented in PyTorch and trained on 4 NVIDIA RTX 4090 GPUs. Specifically, the network is trained end-to-end for 120 epochs using the AdamW optimizer with a batch size of 144 and an initial learning rate of 1e-4. Regarding the optimization, the global trade-off parameter is set to $\lambda_{GCC}=10$ to balance the geometric consistency loss with the retrieval loss. Furthermore, the hierarchical balancing factors for the DS mechanism are set to $\lambda_{deep}=1.0$ and $\lambda_{shallow}=0.5$. All baseline methods are trained and evaluated using the same data split for fair comparison.

\begin{figure*}[!htb]
    \centering
    \begin{tblr}{
        colspec = {ccc},
        colsep = 3pt, 
        rowsep = 0pt, 
        vline{2,3} = {1}{dashed, 1.5pt}
    }
        \includegraphics[height=0.44\textwidth]{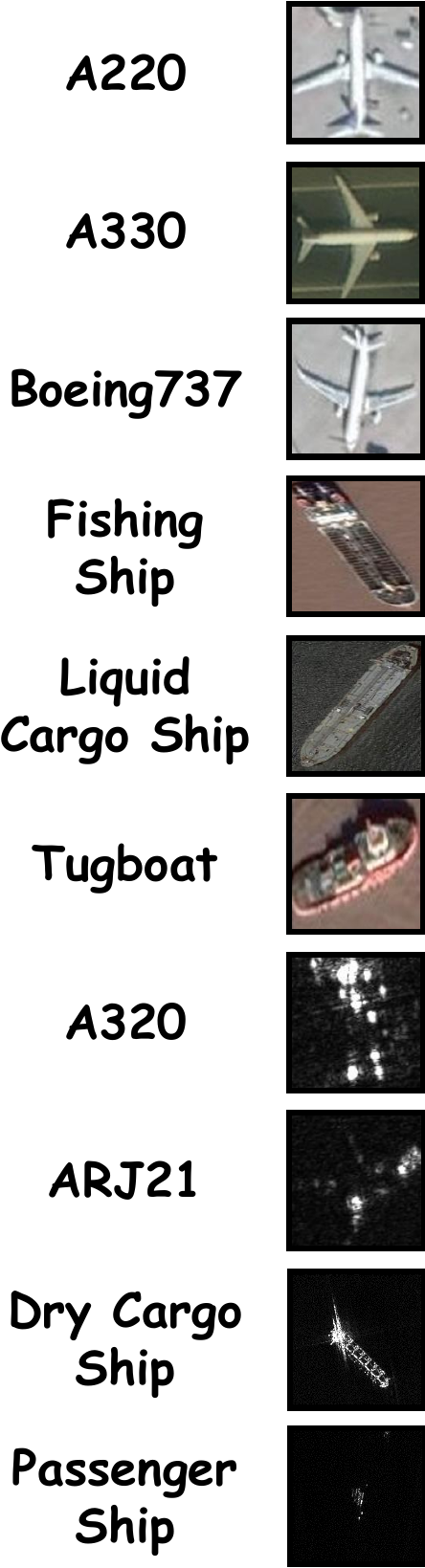} &
        \includegraphics[height=0.44\textwidth]{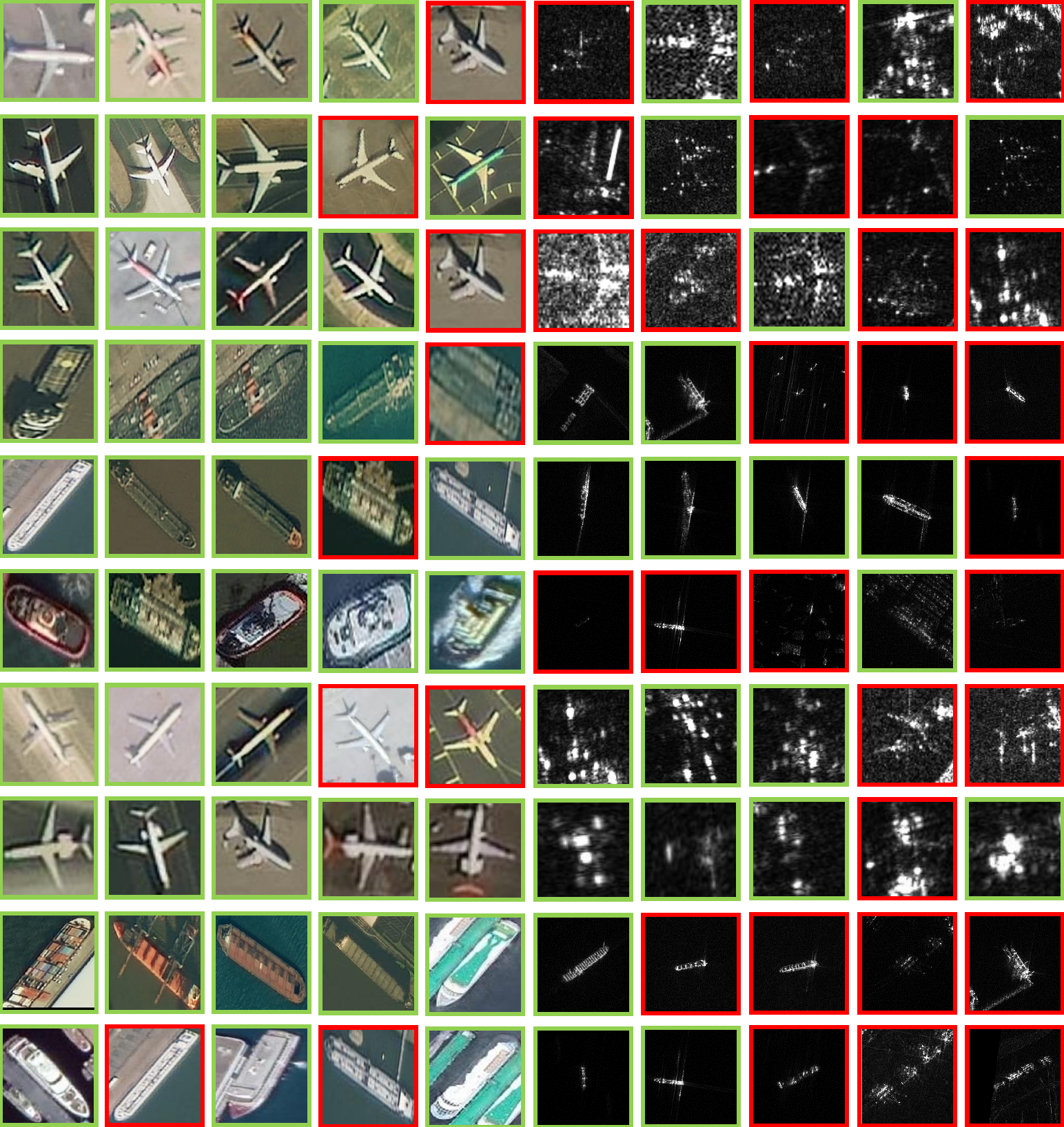} &
        \includegraphics[height=0.44\textwidth]{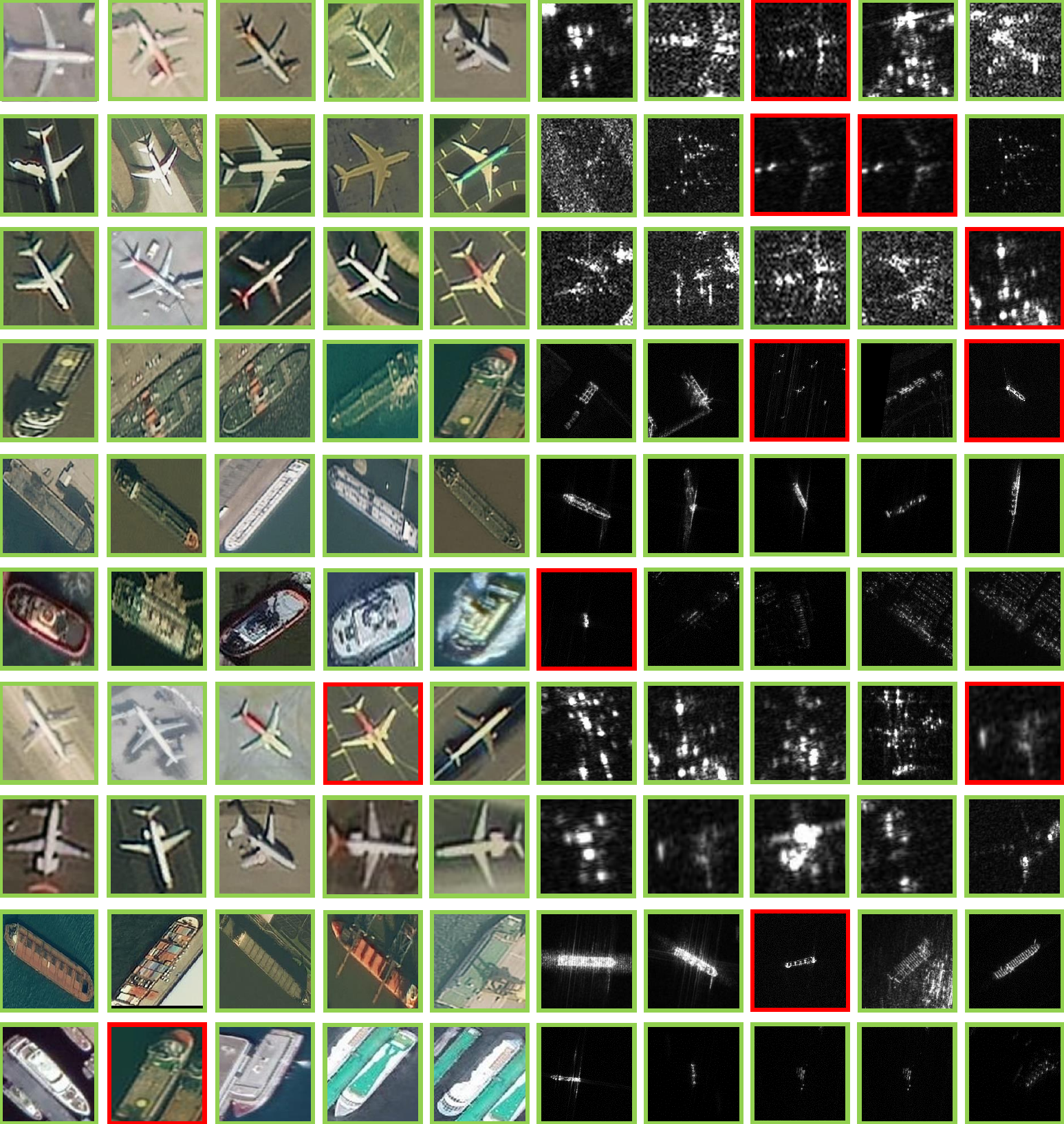} \\
        
        \footnotesize{(a) Query Samples} & 
        \footnotesize{(b) Retrieval Results by TransOSS~\cite{wang2025cross}} & 
        \footnotesize{(c) Retrieval Results by GeoMamba (Ours)}
    \end{tblr}

    \caption{Qualitative comparison of fine-grained retrieval results. Panel (a) shows the query samples, panel (b) displays the retrieval results by TransOSS, and panel (c) presents the results of our proposed GeoMamba. The retrieved results in (b) and (c) are sorted by feature distance from smallest to largest. Correct and incorrect matches are indicated by green and red boxes, respectively. The top six rows correspond to optical queries, while the bottom four rows correspond to SAR queries.}
    \label{fig:compare_vis}
\end{figure*}

\begin{figure*}[!htb]
    \centering
    \setlength{\tabcolsep}{1pt}
    \begin{tabular}{ccccc}
        \includegraphics[height=6.8cm,keepaspectratio]{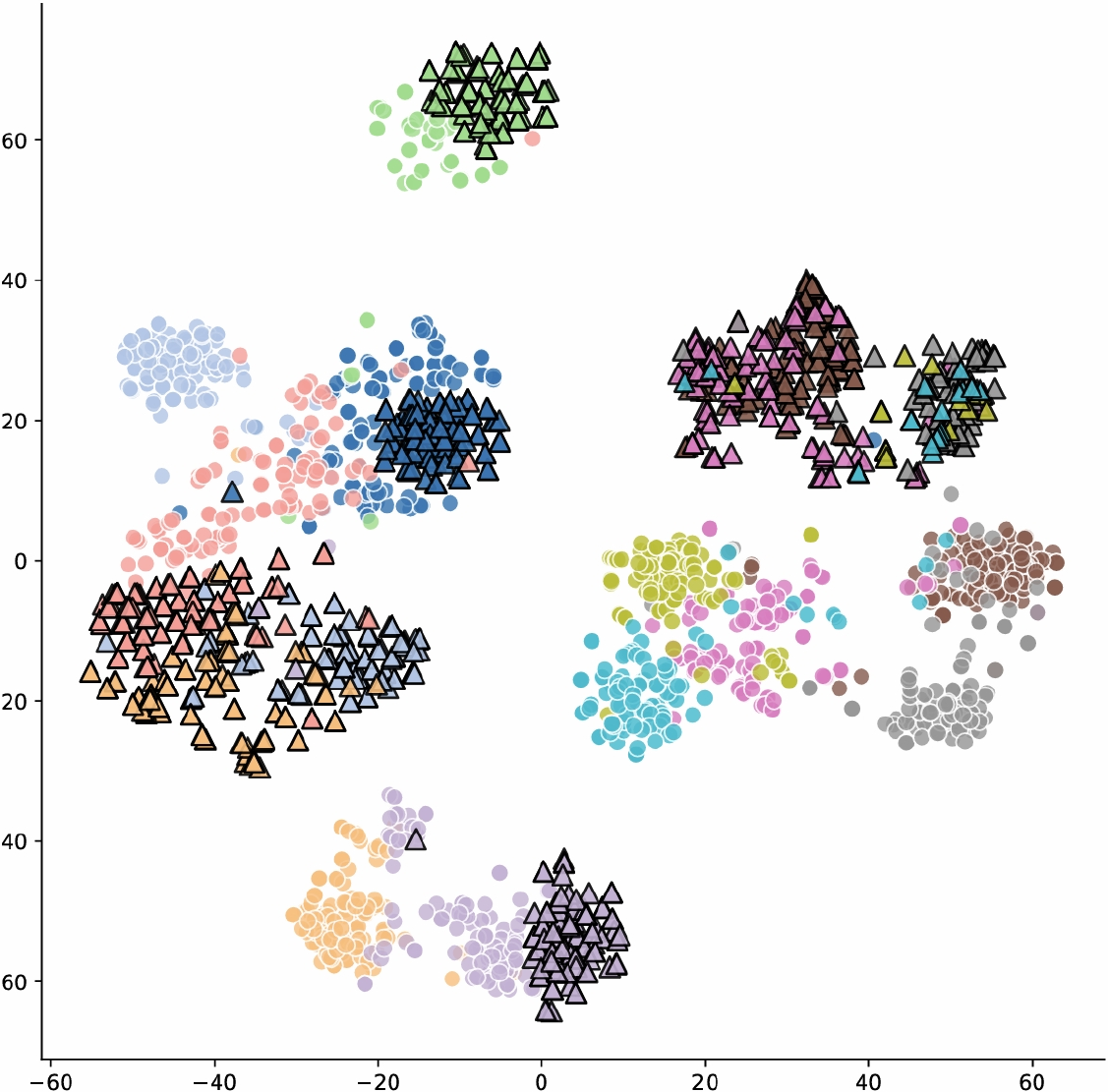} &\quad\quad&
        \includegraphics[height=6.8cm,keepaspectratio]{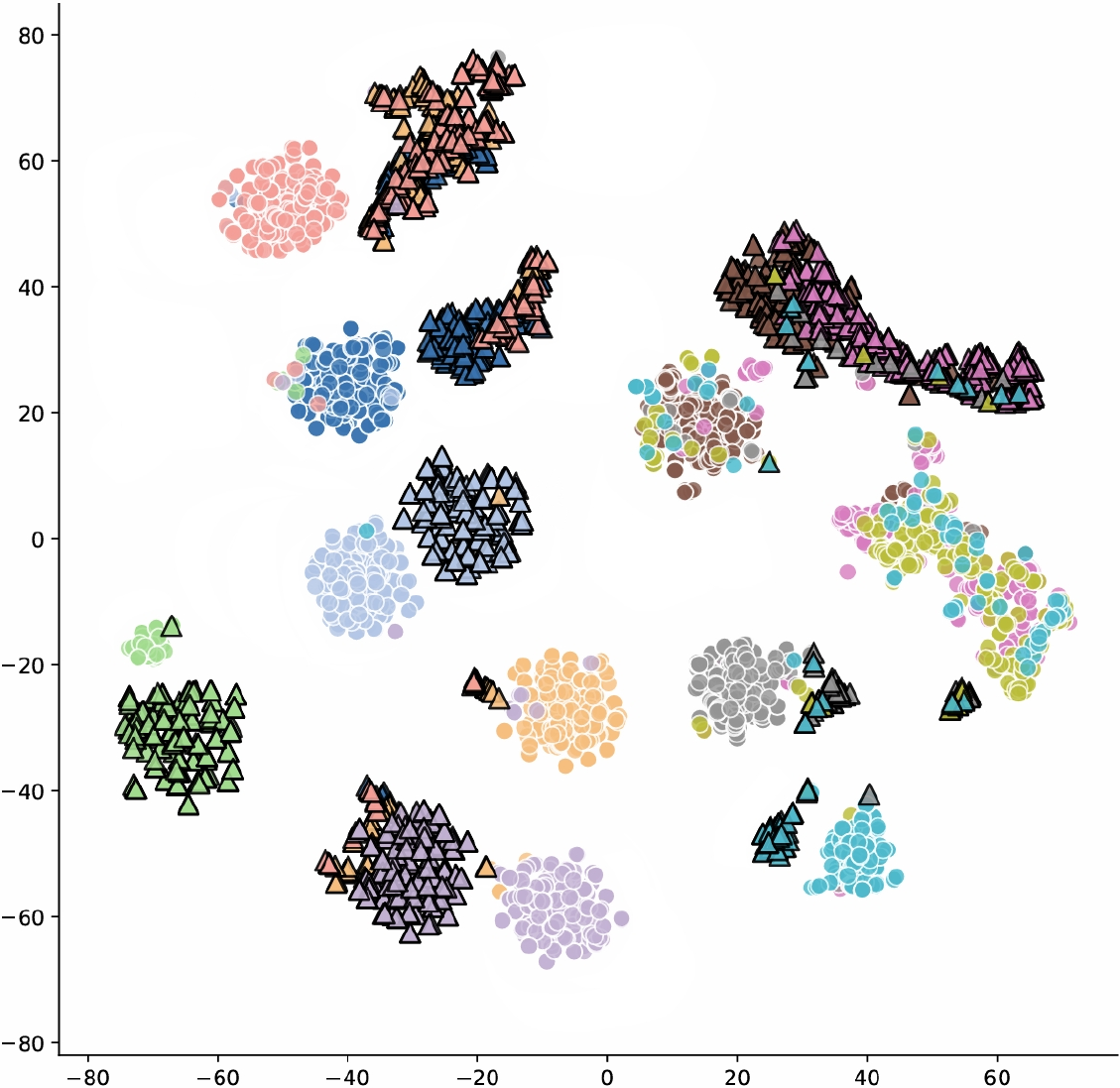} &\quad\quad&
        \includegraphics[height=6.3cm,keepaspectratio]{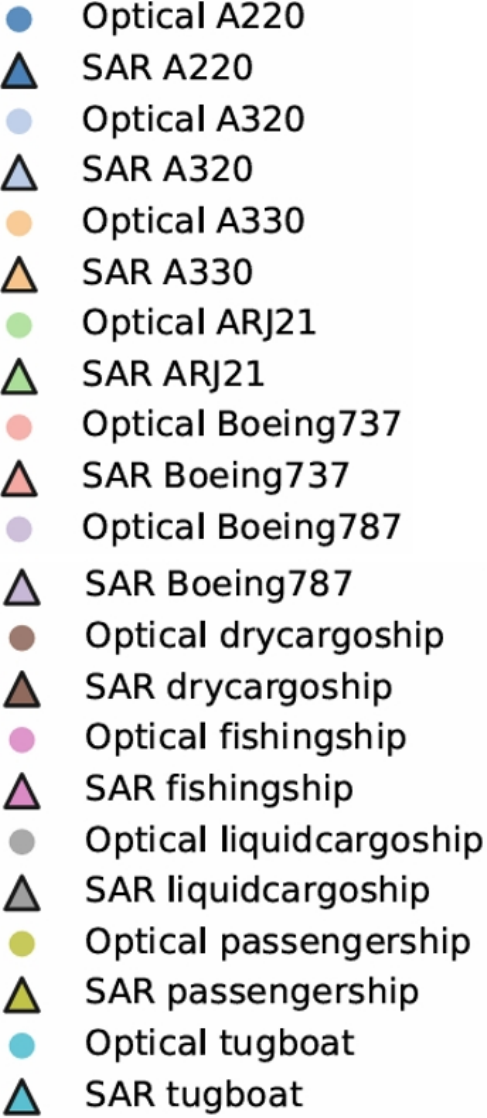} \\
        \footnotesize{(a) TransOSS~\cite{wang2025cross}}&\quad& \footnotesize{(b) GeoMamba (Ours)}&\quad&~
    \end{tabular}

    \caption{t-SNE visualization of the learned feature embeddings on the FGOS-as dataset. The axes represent the 2D projected embedding space. Circles denote optical samples, and black-edged triangles represent SAR samples. Panel (a) displays features extracted by TransOSS, exhibiting a significant modality gap where samples are primarily clustered by distinct modalities. Panel (b) displays features extracted by our proposed GeoMamba, demonstrating that different features belonging to the same fine-grained category are tightly aligned and clustered together.}

    \label{fig:t-SNE}
\end{figure*}

\subsection{Comparisons With State-of-the-Art Methods}
In this section, we present comparison between the proposed GeoMamba and SOTA methods and evaluate the retrieval performance from both quantitative and qualitative aspects.

\subsubsection{Quantitative Analysis} 
As shown in Table \ref{tab:comparison}, GeoMamba consistently achieves the best overall performance across all evaluation settings. Specifically, under the All-to-All setting, GeoMamba achieves 63.3\% mAP and 77.0\% Rank-$1$. Compared with the second-best method, \emph{i.e.}, TransOSS \cite{wang2025cross}, GeoMamba yields a clear improvement of 4.7\% in mAP and 4.9\% in Rank-1. 
In the more challenging Optical-to-SAR retrieval setting, GeoMamba surpasses TransOSS by 4.2\% in mAP and 8.2\% in Rank-1. 
In the SAR-to-Optical retrieval setting, GeoMamba further outperforms TransOSS by 7.4\% in mAP and 3.9\% in Rank-1, indicating stronger robustness to modality discrepancy. 
These results demonstrate that the proposed GeoMamba successfully extracts highly discriminative and modality-invariant representations for cross-modal retrieval.

\subsubsection{Qualitative Analysis}
To provide qualitative evidence of retrieval performance, Fig.~\ref{fig:compare_vis} compares the fine-grained retrieval results of GeoMamba and the strongest baseline, TransOSS \cite{wang2025cross}, across all 11 categories of the FGOS-as dataset. Inherently, distinguishing these fine-grained categories requires capturing subtle structural cues, such as aircraft wing configuration, engine placement, ship hull proportions, and superstructure layouts. Under SAR speckle noise and complex optical backgrounds, TransOSS often produces false matches (red boxes), indicating difficulty in preserving fine-grained structural information. In contrast, GeoMamba retrieves more accurate cross-modal matches in most cases, suggesting stronger robustness to modality discrepancy and background interference. This robust performance can be attributed to the explicit alignment of continuous optical contours with discrete SAR scattering structures, together with the integration of structural priors, which helps bridge the inherent discrete-continuous modality gap between heterogeneous sensors.

\subsubsection{Feature Representations Analysis}
Fig.~\ref{fig:t-SNE} visualizes the learned embeddings using t-SNE, where circles and triangles denote optical and SAR samples, respectively. The TransOSS features in Fig.~\ref{fig:t-SNE}(a) exhibit larger intra-class dispersion and weaker cross-modal clustering. In comparison, Fig.~\ref{fig:t-SNE}(b) shows that the optical and SAR features of the same category from GeoMamba are clustered more closely in a shared embedding space, while different fine-grained classes remain better separated. This observation suggests that the proposed framework learns more discriminative and better aligned cross-modal representations.

\tabcolsep=0.5pt
\begin{figure*}[t]
    \centering
    \resizebox{0.8\textwidth}{!}{
        \begin{tabular}{cccccc}
            \raisebox{-6ex}{\rotatebox{90}{\small Aircraft (Optical)}} & 
            \includegraphics[width=0.16\textwidth, valign=m]{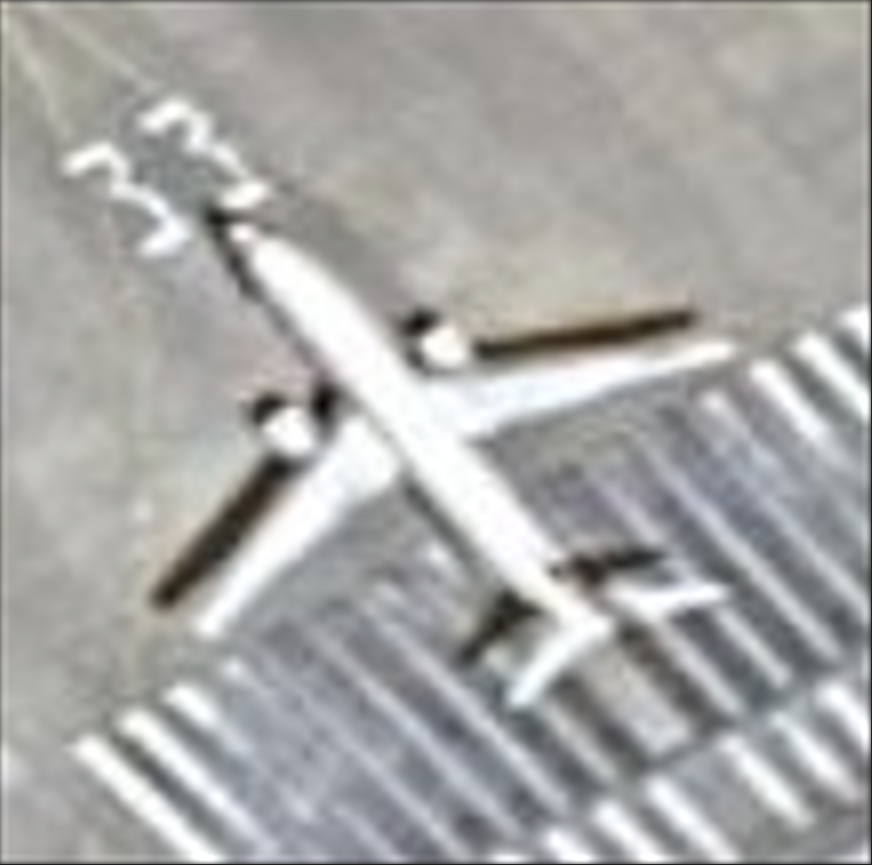} &
            \includegraphics[width=0.16\textwidth, valign=m]{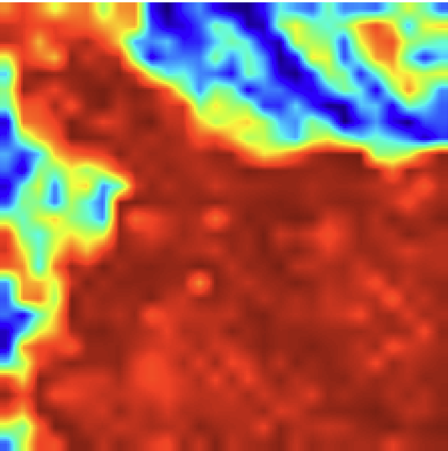} &
            \includegraphics[width=0.16\textwidth, valign=m]{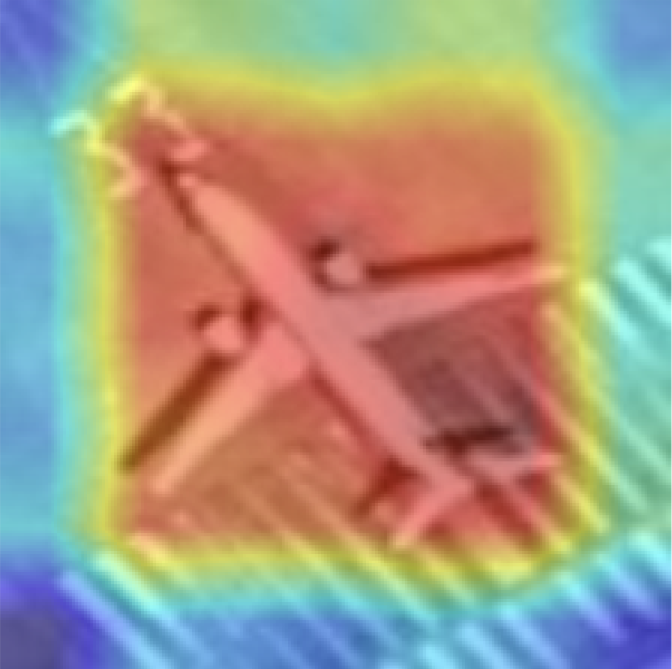} &
            \includegraphics[width=0.16\textwidth, valign=m]{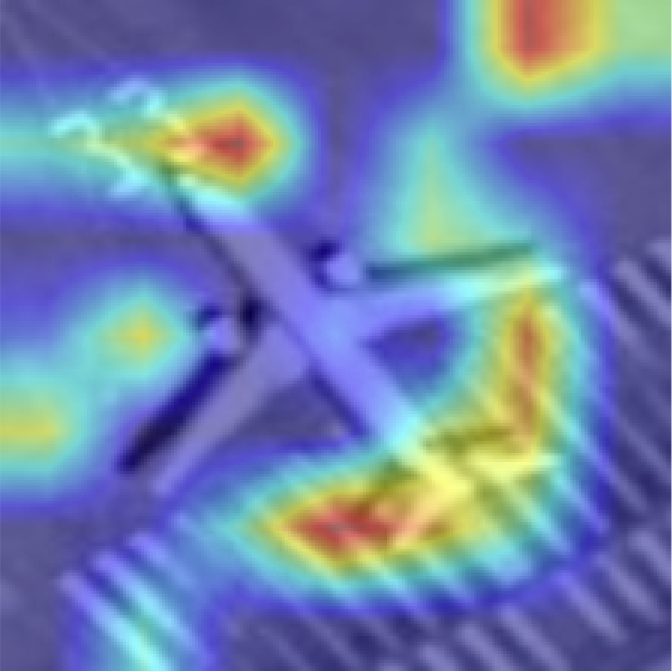} &
            \includegraphics[width=0.16\textwidth, valign=m]{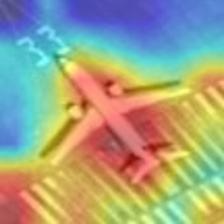} \\
            
            \noalign{\smallskip}
            
            \raisebox{-5ex}{\rotatebox{90}{\small Aircraft (SAR)}} & 
            \includegraphics[width=0.16\textwidth, valign=m]{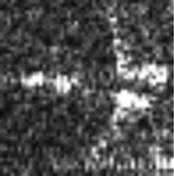} &
            \includegraphics[width=0.16\textwidth, valign=m]{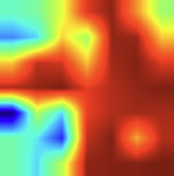} &
            \includegraphics[width=0.16\textwidth, valign=m]{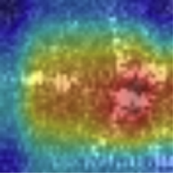} &
            \includegraphics[width=0.16\textwidth, valign=m]{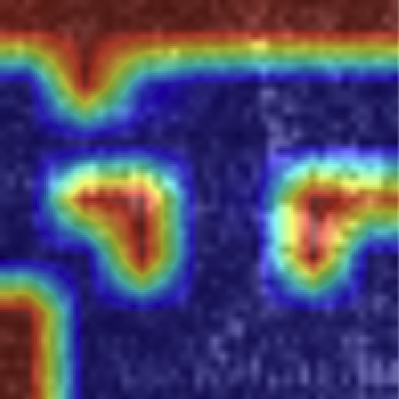} &
            \includegraphics[width=0.16\textwidth, valign=m]{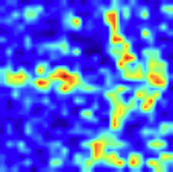} \\

            \noalign{\smallskip}

            \raisebox{-4.5ex}{\rotatebox{90}{\small Ship (Optical)}} & 
            \includegraphics[width=0.16\textwidth, valign=m]{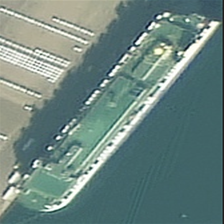} &
            \includegraphics[width=0.16\textwidth, valign=m]{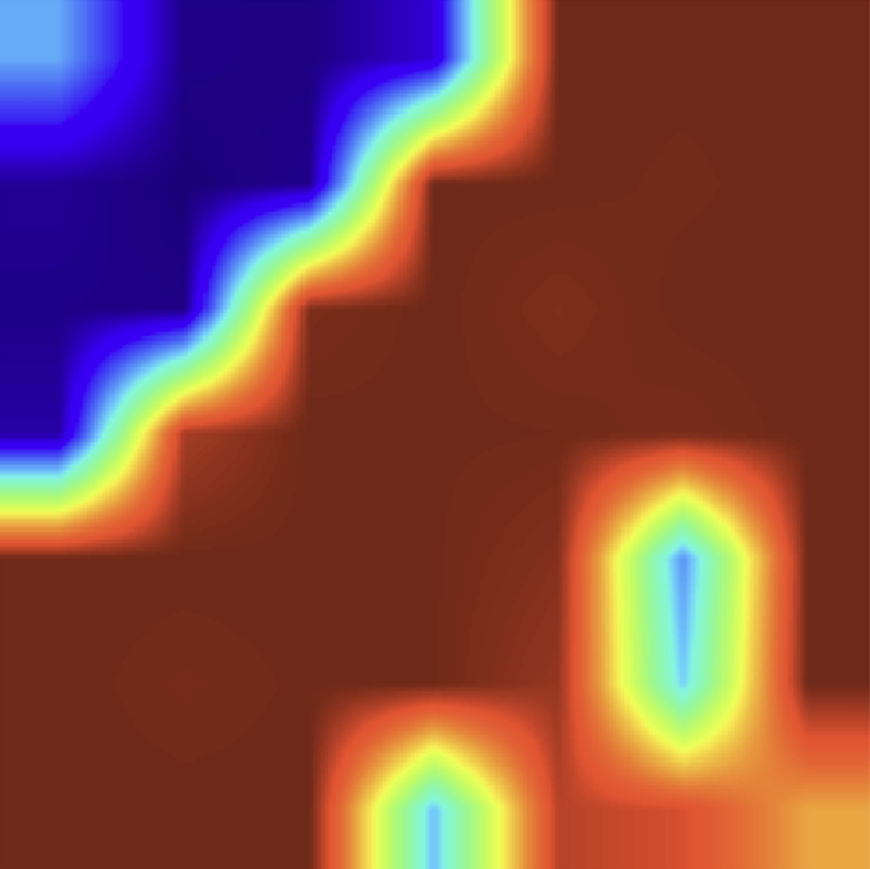} &
            \includegraphics[width=0.16\textwidth, valign=m]{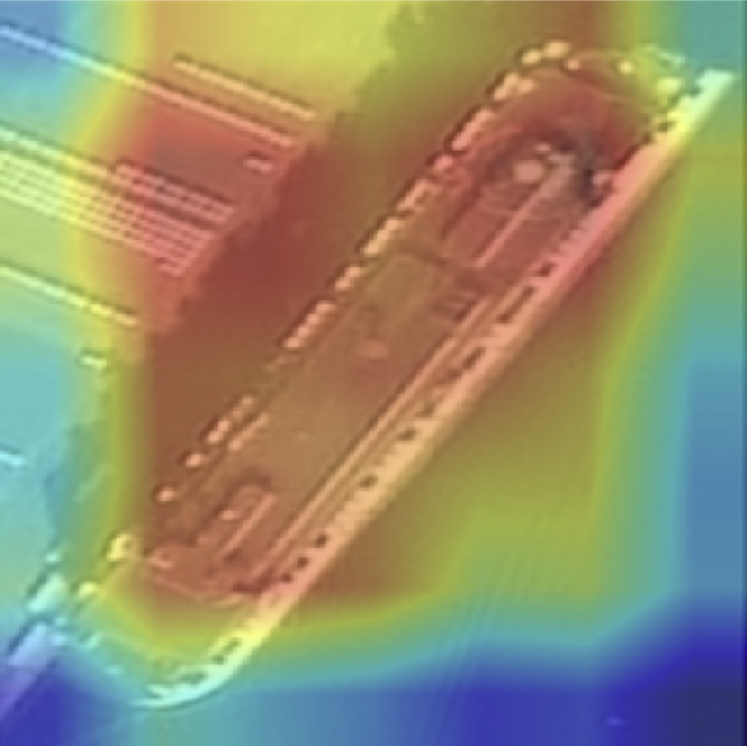} &
            \includegraphics[width=0.16\textwidth, valign=m]{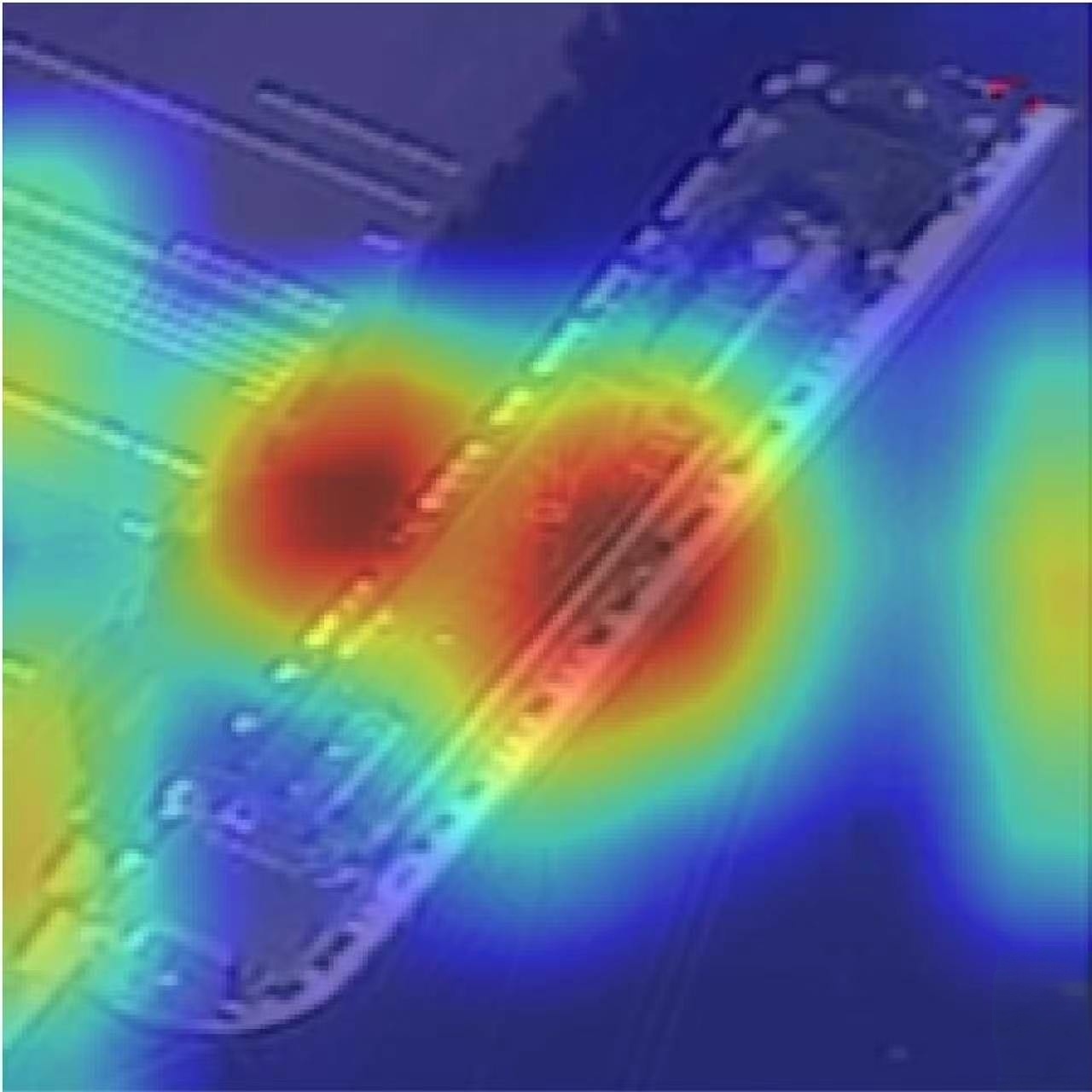} &
            \includegraphics[width=0.16\textwidth, valign=m]{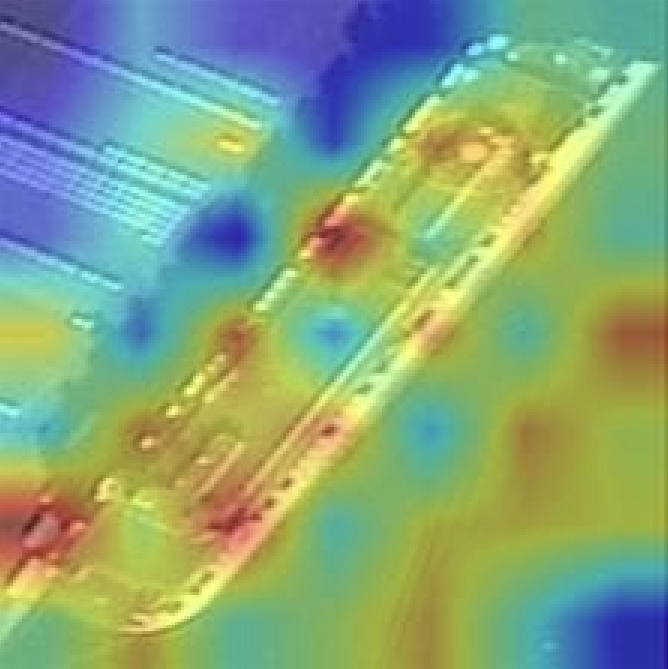} \\

            \noalign{\smallskip}

            \raisebox{-3.5ex}{\rotatebox{90}{\small Ship (SAR)}} & 
            \includegraphics[width=0.16\textwidth, valign=m]{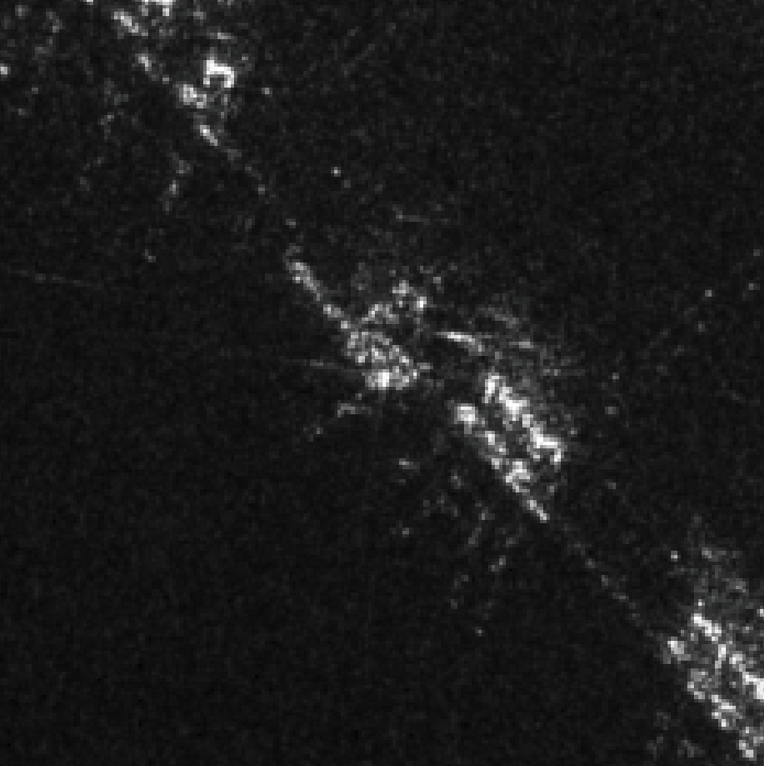} &
            \includegraphics[width=0.16\textwidth, valign=m]{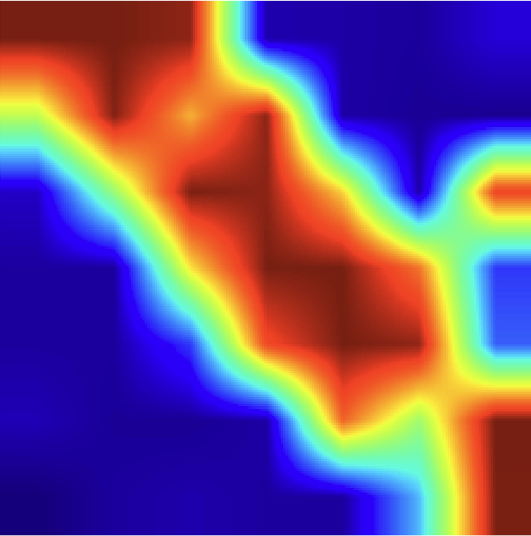} &
            \includegraphics[width=0.16\textwidth, valign=m]{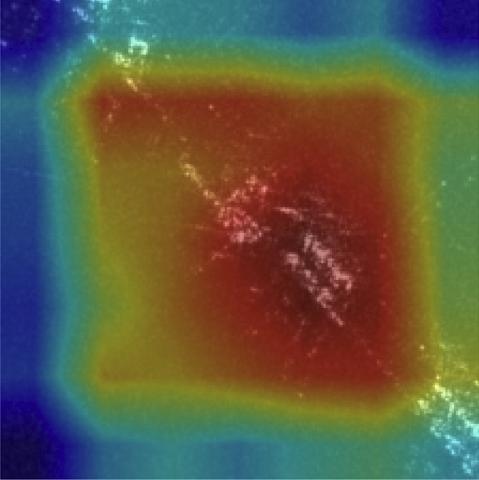} &
            \includegraphics[width=0.16\textwidth, valign=m]{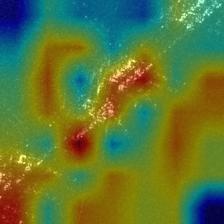} &
            \includegraphics[width=0.16\textwidth, valign=m]{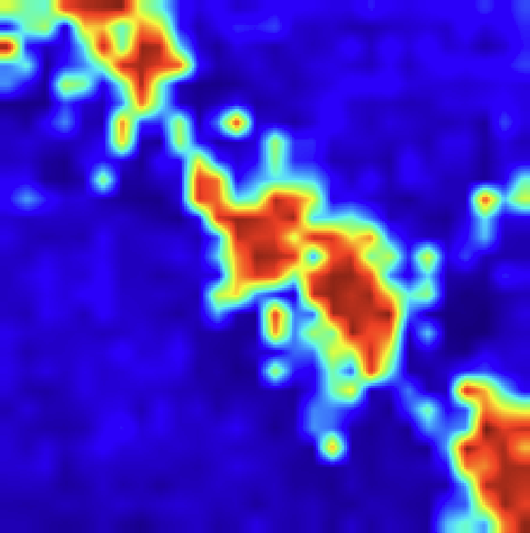} \\
            
            & {\small (a) Raw Input} & {\small (b) Baseline} & {\small (c) Baseline+GFI} & {\small (d) Baseline+GCC} & {\small (e) GeoMamba (Ours)}
        \end{tabular}
    }
    \caption{Visualization of cross-modal feature activation maps for optical and SAR images across different module combinations. The baseline (b) struggles under severe modality discrepancies, mistakenly activating irrelevant background instead of the objects. Integrating the Geometric Feature Injection (GFI) module in (c) begins to refine spatial attention, aligning activation with the targets' intrinsic geometric shapes. Furthermore, incorporating the Geometric Consistency Constraint (GCC) in (d), which intrinsically utilizes a Deep Supervision (DS), captures pristine continuous edges and discrete geometric keypoints to provide a strong structural anchor. Ultimately, our complete GeoMamba framework (e) bridges the inherent modality gap, successfully filtering out complex clutter to remain strictly focused on the objects' invariant geometric signatures.}
    \label{fig:attention_map}
\end{figure*}

\begin{table}[t]
    \centering
    \caption{\textsc{Ablation Study of Key Components. GFI and GCC Denote Geometric Feature Injection and Geometric Consistency Constraint With Structure-Aware Prediction.}}
    \label{table:ablation}
    \renewcommand{\arraystretch}{1.2}
    \setlength{\tabcolsep}{8pt}
    \begin{tabular}{c | c c | c | c}
        \toprule
        Model & GFI & GCC & mAP (\%)  & Change (\%)  \\ 
        \midrule
        (a) & \xmark & \xmark   & 59.6 & -     \\
        (b) & \cmark & \xmark   & 60.8 & \textbf{\textcolor{red}{+ 1.2}} \\
        (c) & \xmark & \cmark   & 61.5 & \textbf{\textcolor{red}{+ 1.9}} \\
        \textbf{Ours} & \cmark & \cmark & \textbf{63.3} & \textbf{\textcolor{red}{+ 3.7}} \\
        \bottomrule
    \end{tabular}
\end{table}

\subsection{Ablation Studies}
To thoroughly evaluate the contributions of the key components in the proposed GeoMamba framework, we conduct comprehensive ablation studies on the FGOS-as benchmark and the results are summarized in Table \ref{table:ablation}. We adopt the dual-stream MambaVision network trained with Identity Loss and Triplet Loss as the baseline model (Model a), while keeping the same training protocol for all variants.

\subsubsection{Effectiveness of Geometric Feature Injection} 
We first examine the impact of the GFI module. As shown by Model (b) in Table \ref{table:ablation}, introducing the interactive feature fusion with auxiliary HRNet priors improves mAP from 59.6\% to 60.8\%, yielding a gain of 1.2\%. This substantial gain explicitly validates our design motivation. Consequently, this cohesive mechanism successfully salvages severely degraded SAR structures and explicitly preserves informative geometric cues such as intrinsic object boundaries and structural contours, which are indispensable for accurate fine-grained retrieval.

\begin{table}[t]
    \centering
    \caption{\textsc{Experimental Results with Different Hyperparameter \texorpdfstring{$\lambda_{GCC}$}{lambda\_GCC} Settings. mAP and Rank-$k$ denote Mean Average Precision and Rank-$k$ Accuracy, respectively.}}
    \label{tab:Hyperparameter}
    \renewcommand{\arraystretch}{1.2}
    \setlength{\tabcolsep}{8pt}
    \begin{tabular}{c  c  c  c  c}
        \toprule
        \texorpdfstring{$\lambda_{GCC}$}{lambda\_GCC} & mAP (\%) & Rank-1 (\%) & Rank-3 (\%) & Rank-5 (\%)\\
        \midrule
        1  & 62.7 & 76.4 & 84.1 & 87.2 \\ 
        5  & 62.1 & 75.9 & 83.4 & 86.8 \\
        \textbf{10} & \textbf{63.3} & \textbf{77.0} & \textbf{84.4} & 87.3 \\
        15 & 62.4 & 76.1 & 83.8 & 87.0 \\
        20 & 62.9 & 76.6 & 84.3 & \textbf{87.4} \\
        \bottomrule
    \end{tabular}
\end{table}

\subsubsection{Effectiveness of Geometric Consistency Constraint}
Building upon the structural priors provided by GFI, we further analyze the GCC strategy. As reported in Table \ref{table:ablation}, adding GCC alone (Model c) improves mAP to 61.5\%, corresponding to a gain of 1.9\% over the baseline. This improvement confirms that explicit geometric supervision provides crucial complementary training signals. Specifically, applying modality-specific operators, notably Harris for discrete geometric keypoints and Sobel for continuous edges, forces the network to explicitly align intrinsic object contours. Furthermore, GCC intrinsically incorporates a Deep Supervision (DS) mechanism to hierarchically anchor these geometric constraints. By enforcing spatial boundaries in the early shallow stages, this structure-aware prediction prevents fine details from being lost during spatial downsampling. This synergistic design effectively bridges the discrete-continuous modality gap and guarantees complete structural retention, ultimately boosting the full model to 63.3\% mAP when combined with GFI.

\subsubsection{Sensitivity Analysis of Hyperparameter \texorpdfstring{$\lambda_{GCC}$}{lambda\_GCC}}
We further investigate the sensitivity of GeoMamba to the geometric consistency weight $\lambda_{GCC}$. As shown in Table~\ref{tab:Hyperparameter}, the retrieval performance does not vary monotonically with $\lambda_{GCC}$, but achieves the best result at $\lambda_{GCC} = 10$ with 63.3\% mAP and 77.0\% Rank-1. Specifically, when $\lambda_{GCC}$ is too small (\emph{e.g.}, 5), the geometric constraint may be insufficient to provide effective cross-modal guidance. In contrast, larger values (\emph{e.g.}, $\lambda_{GCC} \geq 15$) lead to noticeable performance drops, indicating that excessive regularization may interfere with optimization of the primary retrieval objectives. Therefore, $\lambda_{GCC} = 10$ is adopted as the default setting.

\subsubsection{Visual Activation Analysis}
To complement quantitative ablations, Fig.~\ref{fig:attention_map} visualizes step-by-step feature activation maps. Column (b) shows the baseline struggling under severe modality discrepancies, mistakenly activating irrelevant background instead of the actual objects. Integrating the Geometric Feature Injection (GFI) module in (c) begins to refine spatial attention, aligning activation with the targets' intrinsic geometric shapes. Furthermore, incorporating the Geometric Consistency Constraint (GCC) in (d), which intrinsically utilizes a Deep Supervision (DS) mechanism, captures pristine continuous edges and discrete geometric keypoints to provide a strong structural anchor. Ultimately, the complete GeoMamba framework in (e) achieves concentrated responses that precisely trace the modality-invariant geometric signatures of both aircraft and ships. These visual observations corroborate our core methodological claims: driven by GFI and GCC, our network is explicitly forced to focus on intrinsic object contours rather than modality-specific noise.

\section{Conclusion}
\label{sec:Conclusion}
In this paper, we addressed the problem of unaligned cross-modal fine-grained object retrieval for remote sensing objects. To facilitate research in this area, we constructed the FGOS-as dataset, which contains realistic challenges such as modality gap, large-scale variation, and background clutter. We further proposed GeoMamba, a geometry-aware retrieval framework that combines the efficient global modeling capability of the MambaVision backbone with cross-modal structural prior learning. 
By introducing geometric feature guidance and hierarchical consistency supervision, GeoMamba learns more discriminative and better-aligned representations across optical and SAR modalities. Extensive experiments demonstrate that the proposed method achieves state-of-the-art performance on the FGOS-as benchmark.
Future work will extend this framework to zero-shot retrieval and more general multi-modal remote sensing scenarios.

\bibliographystyle{IEEEtran}
\bibliography{references}

\begin{thebibliography}{10}
\providecommand{\url}[1]{#1}
\csname url@samestyle\endcsname
\providecommand{\newblock}{\relax}
\providecommand{\bibinfo}[2]{#2}
\providecommand{\BIBentrySTDinterwordspacing}{\spaceskip=0pt\relax}
\providecommand{\BIBentryALTinterwordstretchfactor}{4}
\providecommand{\BIBentryALTinterwordspacing}{\spaceskip=\fontdimen2\font plus
\BIBentryALTinterwordstretchfactor\fontdimen3\font minus \fontdimen4\font\relax}
\providecommand{\BIBforeignlanguage}[2]{{%
\expandafter\ifx\csname l@#1\endcsname\relax
\typeout{** WARNING: IEEEtran.bst: No hyphenation pattern has been}%
\typeout{** loaded for the language `#1'. Using the pattern for}%
\typeout{** the default language instead.}%
\else
\language=\csname l@#1\endcsname
\fi
#2}}
\providecommand{\BIBdecl}{\relax}
\BIBdecl

\bibitem{tupin2003detection}
F.~Tupin and M.~Roux, ``Detection of building outlines based on the fusion of sar and optical features,'' \emph{ISPRS J. Photogramm. Remote Sens.}, vol.~58, no. 1-2, pp. 71--82, 2003.

\bibitem{wang2025m4}
C.~Wang, W.~S. Lu, X.~M. Li, J.~Yang, and L.~Luo, ``M4-sar: A multi-resolution, multi-polarization, multi-scene, multi-source dataset and benchmark for optical-sar fusion object detection,'' \emph{arXiv preprint arXiv:2505.10931}, 2025.

\bibitem{zhao2025towards}
Z.~Zhao, Y.~Xu, A.~Lu, C.~Li, and J.~Tang, ``Towards robust optical-sar object detection under missing modalities: A dynamic quality-aware fusion framework,'' \emph{arXiv preprint arXiv:2512.22447}, 2025.

\bibitem{zhang2020intelligent}
Y.~Zhang, L.~Guo, Z.~Wang, Y.~Yu, X.~Liu, and F.~Xu, ``Intelligent ship detection in remote sensing images based on multi-layer convolutional feature fusion,'' \emph{Remote Sens.}, vol.~12, no.~20, p. 3316, 2020.

\bibitem{kwak2016disaster}
Y.~Kwak, A.~Yorozuya, and Y.~Iwami, ``Disaster risk reduction using image fusion of optical and sar data before and after tsunami,'' in \emph{Proc. IEEE Aerosp. Conf.}\hskip 1em plus 0.5em minus 0.4em\relax IEEE, 2016, pp. 1--11.

\bibitem{yu2022coastline}
T.~Yu, S.~W. Xu, B.~Y. Tao, and W.~Z. Shao, ``Coastline detection using optical and synthetic aperture radar images,'' \emph{Adv. Space Res.}, vol.~70, no.~1, pp. 70--84, 2022.

\bibitem{zhang2024optical}
Z.~Zhang, L.~Zhang, J.~Wu, and W.~Guo, ``Optical and synthetic aperture radar image fusion for ship detection and recognition: Current state, challenges, and future prospects,'' \emph{IEEE Geosci. Remote Sens. Mag.}, vol.~12, no.~4, pp. 132--168, 2024.

\bibitem{ahmed2025dual}
M.~Ahmed, N.~El-Sheimy, and H.~Leung, ``Dual-modal approach for ship detection: Fusing synthetic aperture radar and optical satellite imagery,'' \emph{Sensors}, vol.~25, no.~2, p. 329, 2025.

\bibitem{rane2025machine}
M.~Rane and S.~Kumar, ``Machine learning based aircraft detection using sar \& optical images,'' \emph{ISPRS Ann. Photogramm. Remote Sens. Spat. Inf. Sci.}, vol.~10, pp. 529--535, 2025.

\bibitem{xiong2020deep}
W.~Xiong, Z.~Xiong, Y.~Zhang, Y.~Cui, and X.~Gu, ``A deep cross-modality hashing network for sar and optical remote sensing images retrieval,'' \emph{IEEE J. Sel. Topics Appl. Earth Observ. Remote Sens.}, vol.~13, pp. 5284--5296, 2020.

\bibitem{sun2021multisensor}
Y.~Sun, S.~Feng, Y.~Ye, X.~Li, J.~Kang, Z.~Huang, and C.~Luo, ``Multisensor fusion and explicit semantic preserving-based deep hashing for cross-modal remote sensing image retrieval,'' \emph{IEEE Trans. Geosci. Remote Sens.}, vol.~60, pp. 1--14, 2021.

\bibitem{huang2024deep}
J.~Huang, Y.~Feng, M.~Zhou, X.~Xiong, Y.~Wang, and B.~Qiang, ``Deep multiscale fine-grained hashing for remote sensing cross-modal retrieval,'' \emph{IEEE Geosci. Remote Sens. Lett.}, vol.~21, pp. 1--5, 2024.

\bibitem{yang2025cross}
J.~Yang and Y.~Tang, ``Cross-modal retrieval algorithm based on patch aggregation,'' in \emph{Proc. 4th Int. Conf. Electron. Inf. Technol. (EIT)}.\hskip 1em plus 0.5em minus 0.4em\relax IEEE, 2025, pp. 632--637.

\bibitem{luo2026dynamic}
Z.~Luo, M.~Meng, and J.~Wu, ``Dynamic patch selection and dual-granularity alignment for cross-modal retrieval,'' \emph{Neurocomputing}, p. 132999, 2026.

\bibitem{CCA}
D.~R. Hardoon, S.~Szedmak, and J.~Shawe-Taylor, ``Canonical correlation analysis: An overview with application to learning methods,'' \emph{Neural Comput.}, vol.~16, no.~12, pp. 2639--2664, 2004.

\bibitem{KCCA}
A.~A. Nielsen, ``Multiset canonical correlations analysis and multispectral, truly multitemporal remote sensing data,'' \emph{IEEE Trans. Image Process.}, vol.~11, no.~3, pp. 293--305, 2002.

\bibitem{LMNN}
K.~Q. Weinberger and L.~K. Saul, ``Distance metric learning for large margin nearest neighbor classification.'' \emph{J. Mach. Learn. Res.}, vol.~10, no.~2, 2009.

\bibitem{CMSSH}
M.~M. Bronstein, A.~M. Bronstein, F.~Michel, and N.~Paragios, ``Data fusion through cross-modality metric learning using similarity-sensitive hashing,'' in \emph{Proc. IEEE CVPR}.\hskip 1em plus 0.5em minus 0.4em\relax IEEE, 2010, pp. 3594--3601.

\bibitem{CVH}
S.~Kumar and R.~Udupa, ``Learning hash functions for cross-view similarity search,'' in \emph{Proc. IJCAI}, vol.~22, no.~1, 2011, p. 1360.

\bibitem{GMA}
A.~Sharma, A.~Kumar, H.~Daume, and D.~W. Jacobs, ``Generalized multiview analysis: A discriminative latent space,'' in \emph{Proc. IEEE CVPR}.\hskip 1em plus 0.5em minus 0.4em\relax IEEE, 2012, pp. 2160--2167.

\bibitem{SIFT}
F.~Dellinger, J.~Delon, Y.~Gousseau, J.~Michel, and F.~Tupin, ``Sar-sift: a sift-like algorithm for sar images,'' \emph{IEEE Trans. Geosci. Remote Sens.}, vol.~53, no.~1, pp. 453--466, 2014.

\bibitem{LBP}
B.~Demir and L.~Bruzzone, ``A novel active learning method in relevance feedback for content-based remote sensing image retrieval,'' \emph{IEEE Trans. Geosci. Remote Sens.}, vol.~53, no.~5, pp. 2323--2334, 2014.

\bibitem{liu2020parameter}
H.~Liu, X.~Tan, and X.~Zhou, ``Parameter sharing exploration and hetero-center triplet loss for visible-thermal person re-identification,'' \emph{IEEE Trans. Multimedia}, vol.~23, pp. 4414--4425, 2020.

\bibitem{liang2024bridging}
T.~Liang, Y.~Jin, W.~Liu, T.~Wang, S.~Feng, and Y.~Li, ``Bridging the gap: Multi-level cross-modality joint alignment for visible-infrared person re-identification,'' \emph{IEEE Trans. Circuits Syst. Video Technol.}, vol.~34, no.~8, pp. 7683--7698, 2024.

\bibitem{park2021learning}
H.~Park, S.~Lee, J.~Lee, and B.~Ham, ``Learning by aligning: Visible-infrared person re-identification using cross-modal correspondences,'' in \emph{Proc. IEEE/CVF ICCV}, 2021, pp. 12\,046--12\,055.

\bibitem{ye2021deep}
M.~Ye, J.~Shen, G.~Lin, T.~Xiang, L.~Shao, and S.~C. Hoi, ``Deep learning for person re-identification: A survey and outlook,'' \emph{IEEE Trans. Pattern Anal. Mach. Intell.}, vol.~44, no.~6, pp. 2872--2893, 2021.

\bibitem{dosovitskiy2020image}
A.~Dosovitskiy, ``An image is worth 16x16 words: Transformers for image recognition at scale,'' \emph{arXiv preprint arXiv:2010.11929}, 2020.

\bibitem{touvron2021training}
H.~Touvron, M.~Cord, M.~Douze, F.~Massa, A.~Sablayrolles, and H.~J{\'e}gou, ``Training data-efficient image transformers \& distillation through attention,'' in \emph{Proc. ICML}.\hskip 1em plus 0.5em minus 0.4em\relax PMLR, 2021, pp. 10\,347--10\,357.

\bibitem{he2021transreid}
S.~He, H.~Luo, P.~Wang, F.~Wang, H.~Li, and W.~Jiang, ``Transreid: Transformer-based object re-identification,'' in \emph{Proc. IEEE/CVF ICCV}, 2021, pp. 15\,013--15\,022.

\bibitem{zheng2024versatile}
W.-S. Zheng, J.~Yan, and Y.-X. Peng, ``A versatile framework for multi-scene person re-identification,'' \emph{IEEE Trans. Pattern Anal. Mach. Intell.}, vol.~47, no.~3, pp. 1362--1380, 2024.

\bibitem{chen2023beyond}
W.~Chen, X.~Xu, J.~Jia, H.~Luo, Y.~Wang, F.~Wang, R.~Jin, and X.~Sun, ``Beyond appearance: a semantic controllable self-supervised learning framework for human-centric visual tasks,'' in \emph{Proc. IEEE/CVF CVPR}, 2023, pp. 15\,050--15\,061.

\bibitem{wang2025cross}
H.~Wang, S.~Li, J.~Yang, Y.~Liu, Y.~Lv, and Z.~Zhou, ``Cross-modal ship re-identification via optical and sar imagery: A novel dataset and method,'' in \emph{Proc. IEEE/CVF ICCV}, October 2025, pp. 7873--7883.

\bibitem{gu2021efficiently}
A.~Gu, K.~Goel, and C.~R{\'e}, ``Efficiently modeling long sequences with structured state spaces,'' \emph{arXiv preprint arXiv:2111.00396}, 2021.

\bibitem{nguyen2022s4nd}
E.~Nguyen, K.~Goel, A.~Gu, G.~Downs, P.~Shah, T.~Dao, S.~Baccus, and C.~R{\'e}, ``S4nd: Modeling images and videos as multidimensional signals with state spaces,'' \emph{Adv. Neural Inf. Process. Syst. (NeurIPS)}, vol.~35, pp. 2846--2861, 2022.

\bibitem{zhu2024vision}
L.~Zhu, B.~Liao, Q.~Zhang, X.~Wang, W.~Liu, and X.~Wang, ``Vision mamba: Efficient visual representation learning with bidirectional state space model,'' \emph{arXiv preprint arXiv:2401.09417}, 2024.

\bibitem{liu2024vmamba}
Y.~Liu, Y.~Tian, Y.~Zhao, H.~Yu, L.~Xie, Y.~Wang, Q.~Ye, J.~Jiao, and Y.~Liu, ``Vmamba: Visual state space model,'' \emph{Adv. Neural Inf. Process. Syst. (NeurIPS)}, vol.~37, pp. 103\,031--103\,063, 2024.

\bibitem{2025mambavision}
A.~Hatamizadeh and J.~Kautz, ``Mambavision: A hybrid mamba-transformer vision backbone,'' in \emph{Proc. IEEE/CVF CVPR}, 2025, pp. 25\,261--25\,270.

\bibitem{MambaHSI}
Y.~Li, Y.~Luo, L.~Zhang, Z.~Wang, and B.~Du, ``Mambahsi: Spatial–spectral mamba for hyperspectral image classification,'' \emph{IEEE Trans. Geosci. Remote Sens.}, vol.~62, pp. 1--16, 2024.

\bibitem{zhu2024samba}
Q.~Zhu, Y.~Cai, Y.~b. Fang, Y.~Yang, C.~Chen, L.~Fan, and A.~Nguyen, ``Samba: Semantic segmentation of remotely sensed images with state space model,'' \emph{Heliyon}, vol.~10, no.~19, 2024.

\bibitem{chen2024rsmamba}
K.~Chen, B.~Chen, C.~Liu, W.~Li, Z.~Zou, and Z.~Shi, ``Rsmamba: Remote sensing image classification with state space model,'' \emph{IEEE Geosci. Remote Sens. Lett.}, vol.~21, pp. 1--5, 2024.

\bibitem{ye}
Y.~Ye, J.~Shan, L.~Bruzzone, and L.~Shen, ``Robust registration of multimodal remote sensing images based on structural similarity,'' \emph{IEEE Trans. Geosci. Remote Sens.}, vol.~55, no.~5, pp. 2941--2958, 2017.

\bibitem{ye2017robust}
Y.~Ye, L.~Shen, M.~Hao, J.~Wang, and Z.~Xu, ``Robust optical-to-sar image matching based on shape properties,'' \emph{IEEE Geosci. Remote Sens. Lett.}, vol.~14, no.~4, pp. 564--568, 2017.

\bibitem{wang2023new}
Z.~Wang, Z.~Huang, and M.~Datcu, ``A new perspective on physics guided learning for sar image interpretation,'' in \emph{Proc. IEEE IGARSS}.\hskip 1em plus 0.5em minus 0.4em\relax IEEE, 2023, pp. 1926--1929.

\bibitem{yi2024deep}
L.~Yi, D.~Lan, Z.~Ke’er, and D.~Yuang, ``Deep network for sar target recognition based on attribute scattering center convolutional kernel modulation,'' \emph{J. Radars}, vol.~13, no.~2, pp. 443--456, 2024.

\bibitem{xiong2025sar}
X.~Xiong, X.~Zhang, W.~Jiang, L.~Liu, Y.~Liu, and T.~Liu, ``Sar-gtr: Attributed scattering information guided sar graph transformer recognition algorithm,'' \emph{arXiv preprint arXiv:2505.08547}, 2025.

\bibitem{yang2025deep}
Y.~Yang and H.~Zhao, ``Deep learning-based sar target recognition: A dual-perspective survey of closed set and open set.'' \emph{Appl. Sci.}, vol.~15, no.~23, 2025.

\bibitem{8510891}
D.~Li, Z.~Zhang, X.~Chen, and K.~Huang, ``A richly annotated pedestrian dataset for person retrieval in real surveillance scenarios,'' \emph{IEEE Trans. Image Process.}, vol.~28, no.~4, pp. 1575--1590, 2019.

\bibitem{di2021public}
Y.~Di, Z.~Jiang, and H.~Zhang, ``A public dataset for fine-grained ship classification in optical remote sensing images,'' \emph{Remote Sens.}, vol.~13, no.~4, p. 747, 2021.

\bibitem{yuming2025osdataset2}
Y.~Xiang, J.~Chen, Z.~Hong, N.~Jiao, F.~Wang, H.~You, and X.~Tong, ``Osdataset2. 0: Sar-optical image matching dataset and evaluation benchmark,'' \emph{J. Radars}, vol.~14, pp. 1--13, 2025.

\bibitem{yang2010bag}
Y.~Yang and S.~Newsam, ``Bag-of-visual-words and spatial extensions for land-use classification,'' in \emph{Proc. 18th SIGSPATIAL Int. Conf. Adv. Geogr. Inf. Syst.}, 2010, pp. 270--279.

\bibitem{xia2010structural}
G.-S. Xia, W.~Yang, J.~Delon, Y.~Gousseau, H.~Sun, and H.~Ma{\^\i}tre, ``Structural high-resolution satellite image indexing,'' in \emph{Proc. ISPRS TC VII Symp.}, vol.~38, 2010, pp. 298--303.

\bibitem{li2018learning}
Y.~Li, Y.~Zhang, X.~Huang, and J.~Ma, ``Learning source-invariant deep hashing convolutional neural networks for cross-source remote sensing image retrieval,'' \emph{IEEE Trans. Geosci. Remote Sens.}, vol.~56, no.~11, pp. 6521--6536, 2018.

\bibitem{sun2021deep}
Y.~Sun, S.~Feng, Y.~Ye, X.~Li, and J.~Kang, ``A deep cross-modal hashing technique for large-scale sar and vhr image retrieval,'' in \emph{Proc. BIGSARDATA}.\hskip 1em plus 0.5em minus 0.4em\relax IEEE, 2021, pp. 1--4.

\bibitem{xiong2022interpretable}
W.~Xiong, Z.~Xiong, Y.~Cui, L.~Huang, and R.~Yang, ``An interpretable fusion siamese network for multi-modality remote sensing ship image retrieval,'' \emph{IEEE Trans. Circuits Syst. Video Technol.}, vol.~33, no.~6, pp. 2696--2712, 2022.

\bibitem{xu2023sar}
W.~Xu, X.~Yuan, Q.~Hu, and J.~Li, ``Sar-optical feature matching: A large-scale patch dataset and a deep local descriptor,'' \emph{Int. J. Appl. Earth Obs. Geoinf.}, vol. 122, p. 103433, 2023.

\bibitem{sun2022fair1m}
X.~Sun, P.~Wang, Z.~Yan, F.~Xu, R.~Wang, W.~Diao, J.~Chen, J.~Li, Y.~Feng, T.~Xu \emph{et~al.}, ``Fair1m: A benchmark dataset for fine-grained object recognition in high-resolution remote sensing imagery,'' \emph{ISPRS J. Photogramm. Remote Sens.}, vol. 184, pp. 116--130, 2022.

\bibitem{hou2020fusar}
X.~Hou, W.~Ao, Q.~Song \emph{et~al.}, ``{FUSAR-Ship}: Building a high-resolution {SAR}-{AIS} matchup dataset of {Gaofen-3} for ship detection and recognition,'' \emph{Sci. China Inf. Sci.}, vol.~63, no.~4, p. 140303, 2020.

\bibitem{zhirui2023sar}
W.~Zhirui, K.~Yuzhuo, Z.~Xuan, W.~Yuelei, Z.~Ting, and S.~Xian, ``Sar-aircraft-1.0: High-resolution sar aircraft detection and recognition dataset,'' \emph{J. Radars}, vol.~12, no.~4, pp. 906--922, 2023.

\bibitem{xian2019air}
S.~Xian, W.~Zhirui, S.~Yuanrui, D.~Wenhui, Z.~Yue, and F.~Kun, ``Air-sarship-1.0: High-resolution sar ship detection dataset,'' \emph{J. Radars}, vol.~8, no.~6, pp. 852--863, 2019.

\bibitem{10806766}
Y.~Wu, Y.~Suo, Q.~Meng, W.~Dai, T.~Miao, W.~Zhao, Z.~Yan, W.~Diao, G.~Xie, Q.~Ke, Y.~Zhao, K.~Fu, and X.~Sun, ``Fair-csar: A benchmark dataset for fine-grained object detection and recognition based on single-look complex sar images,'' \emph{IEEE Trans. Geosci. Remote Sens.}, vol.~63, pp. 1--22, 2025.

\bibitem{chen2020gaussian}
B.-H. Chen, Y.-S. Tseng, and J.-L. Yin, ``Gaussian-adaptive bilateral filter,'' \emph{IEEE Signal Process. Lett.}, vol.~27, pp. 1670--1674, 2020.

\bibitem{ma2014optimized}
T.~Ma, L.~Li, S.~Ji, X.~Wang, Y.~Tian, A.~Al-Dhelaan, and M.~Al-Rodhaan, ``Optimized laplacian image sharpening algorithm based on graphic processing unit,'' \emph{Physica A}, vol. 416, pp. 400--410, 2014.

\bibitem{qiu2004speckle}
F.~Qiu, J.~Berglund, J.~R. Jensen, P.~Thakkar, and D.~Ren, ``Speckle noise reduction in sar imagery using a local adaptive median filter,'' \emph{GIScience Remote Sens.}, vol.~41, no.~3, pp. 244--266, 2004.

\bibitem{mohammed2025contrast}
I.~M. Mohammed and N.~A.~M. Isa, ``Contrast limited adaptive local histogram equalization method for poor contrast image enhancement,'' \emph{IEEE Access}, 2025.

\bibitem{huang2025physics}
Z.~Huang, L.~Liu, S.~M. Yang, Z.~Wang, G.~Cheng, and J.~Han, ``Physics-guided detector for sar airplanes,'' \emph{IEEE Trans. Circuits Syst. Video Technol.}, 2025.

\bibitem{vaswani2017attention}
A.~Vaswani, N.~Shazeer, N.~Parmar, J.~Uszkoreit, L.~Jones, A.~N. Gomez, {\L}.~Kaiser, and I.~Polosukhin, ``Attention is all you need,'' \emph{Adv. Neural Inf. Process. Syst. (NeurIPS)}, vol.~30, 2017.

\bibitem{Zhang_2023_CVPR}
Y.~Zhang and H.~Wang, ``Diverse embedding expansion network and low-light cross-modality benchmark for visible-infrared person re-identification,'' in \emph{Proc. IEEE/CVF CVPR}, June 2023, pp. 2153--2162.

\bibitem{liu2025advancing}
B.~Liu, R.~Huang, X.~Pan, C.~Li, J.~Sun, J.~Dong, and X.~Wang, ``Advancing ship re-identification in the wild: The shipreid-2400 benchmark dataset and d2internet baseline method,'' in \emph{Proc. 48th Int. ACM SIGIR Conf. Res. Develop. Inf. Retr.}, 2025, pp. 106--115.

\bibitem{deng2009imagenet}
J.~Deng, W.~Dong, R.~Socher, L.-J. Li, K.~Li, and L.~Fei-Fei, ``Imagenet: A large-scale hierarchical image database,'' in \emph{Proc. IEEE CVPR}, 2009, pp. 248--255.

\end{thebibliography}

\vfill

\end{document}